\documentclass[10pt,twocolumn,letterpaper]{article}

\usepackage{cvpr}              % To produce the CAMERA-READY version
\usepackage{graphicx}
\usepackage{arydshln}
\usepackage{float}
\usepackage{pifont}
\usepackage[ruled,vlined]{algorithm2e}
\usepackage{lipsum} % just for position
\usepackage{amsmath}
\usepackage{xcolor}
\usepackage{wrapfig}
\usepackage[most]{tcolorbox}
\usepackage{enumitem}
\usepackage{fontawesome}
\usepackage{tcolorbox}
\tcbuselibrary{skins,breakable}
\usepackage{titletoc}
\definecolor{cvprblue}{rgb}{0.21,0.49,0.74}
\usepackage[pagebackref,breaklinks,colorlinks,allcolors=cvprblue]{hyperref}

\def\confName{CVPR}
\def\confYear{2026}

\newcommand{\revise}[1]{\textcolor{black}{#1}}
\newcommand{\final}[1]{\textcolor{black}{#1}}
\definecolor{okgreen}{RGB}{0,150,0}
\definecolor{badred}{RGB}{200,30,40}
\newcommand{\ok}{\textcolor{okgreen}{\ding{51}}}
\newcommand{\bad}{\textcolor{badred}{\ding{55}}}

\newcommand{\dataset}{\textsc{ReasonMap}\xspace}
\definecolor{map_red}{RGB}{239,99,75}
\definecolor{map_blue}{RGB}{99,113,250}
\definecolor{map_green}{RGB}{0,180,139}
\definecolor{map_yellow}{RGB}{229,157,35}
\definecolor{map_gray}{RGB}{165,165,165}
\definecolor{map_purple}{RGB}{128,0,128}

\definecolor{link}{RGB}{147,112,219}

\definecolor{mygreen}{RGB}{93,173,85}
\definecolor{myred}{RGB}{192,57,43}
\newcommand{\resup}[2]{%
  #1 {\fontsize{7.5pt}{1em}\selectfont\textcolor{mygreen}{$\!\uparrow\!$ \textbf{#2}}}%
}
\newcommand{\resdown}[2]{%
  #1 {\fontsize{7.5pt}{1em}\selectfont\textcolor{myred}{$\!\downarrow\!$ \textbf{#2}}}%
}

\title{\dataset: Towards Fine-Grained Visual Reasoning from Transit Maps}

\author{
Sicheng Feng$^{1,2,\dagger}$,~ 
Song Wang$^{3,2,\dagger}$,~ 
Shuyi Ouyang$^{3,2}$,~ 
Lingdong Kong$^2$,~ 
Zikai Song$^{4,2}$,\\ 
\ Jianke Zhu$^3$,~ Huan Wang$^{1,*}$,~ 
Xinchao Wang$^2$
\\[1ex]
$^1$Westlake University \quad $^2$National University of Singapore \quad $^3$Zhejiang University
\\
$^4$Huazhong University of Science and Technology
\\[1ex]
\faGithubAlt~\textbf{Dataset \& Toolkit:} \href{https://fscdc.github.io/ReasonMap}{\texttt{https://fscdc.github.io/ReasonMap}}
\\[1ex]
$^\dagger$Equal contribution. $^*$Corresponding author: \url{wanghuan@westlake.edu.cn}
}

\begin{document}
\twocolumn[{
\maketitle\centering
\captionsetup{type=figure}
\vspace{-0.4cm}
\includegraphics[width=0.99\textwidth]{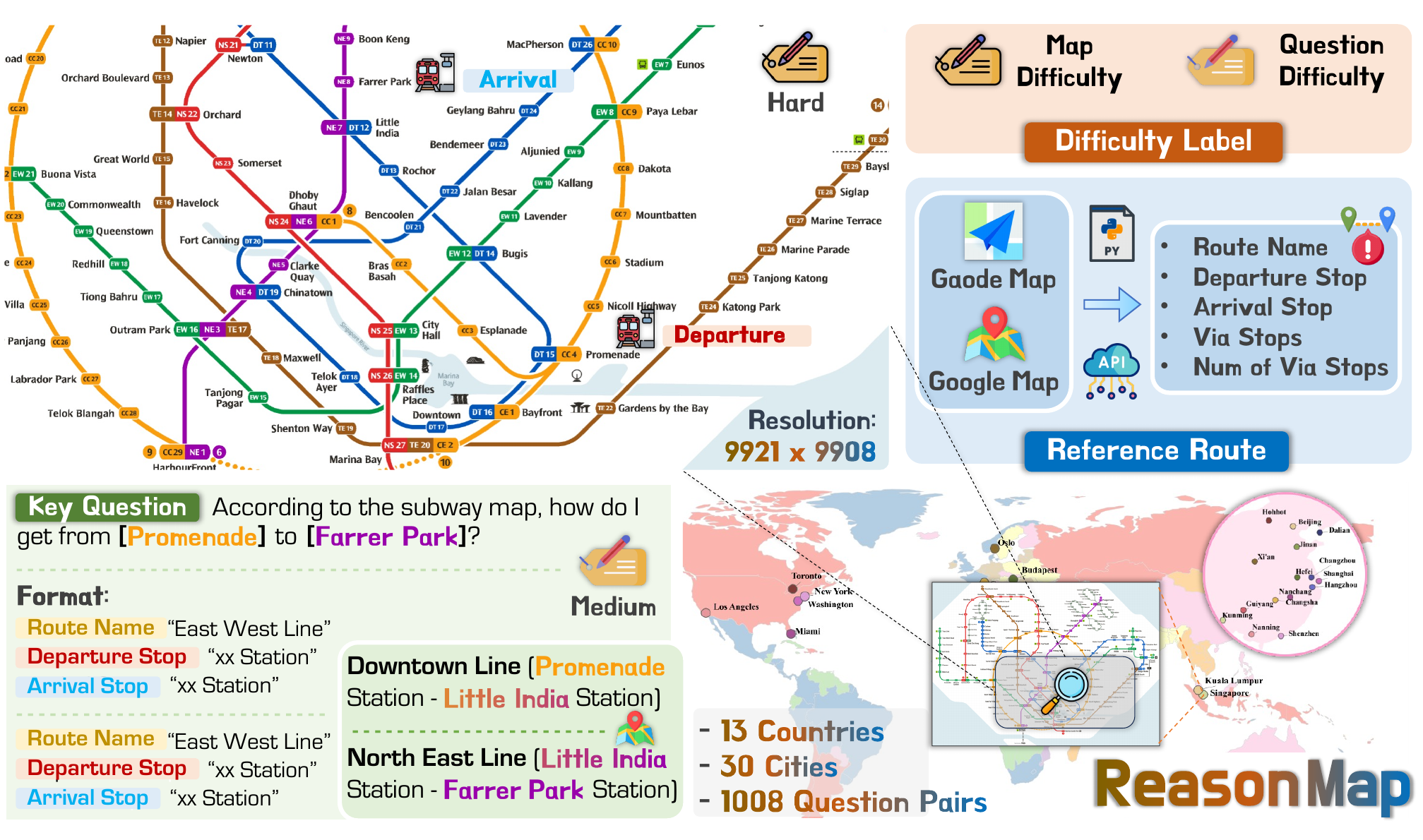}
\vspace{-0.4cm}
\captionof{figure}{
Overview of \dataset. We present a novel benchmark tailored for evaluating the \textbf{fine-grained visual reasoning} capabilities of MLLMs. The dataset comprises $1{,}008$ question-answer pairs derived from high-resolution transit maps across $30$ cities in $13$ countries, featuring diversely structured questions. Further details on the dataset construction pipeline are provided in Section~\ref{sec:dataset}.}
\label{fig:overview}
\vspace{3.5mm}
}]

\maketitle

\begin{abstract}
Multimodal large language models (MLLMs) have demonstrated significant progress in semantic scene understanding and text-image alignment, with reasoning variants enhancing performance on more complex tasks involving mathematics and logic. 
% However, their capacity for reasoning tasks involving fine-grained visual understanding remains insufficiently evaluated.
However, their proficiency in tasks requiring both fine-grained visual understanding and spatial reasoning remains underexplored.
To bridge this gap, we introduce \dataset, a novel benchmark specifically designed to evaluate these capabilities. 
\dataset encompasses high-resolution transit maps from $30$ cities and includes $1{,}008$ question-answer pairs spanning two question types and three templates. 
Furthermore, we design a two-level evaluation pipeline that properly assesses answer correctness and quality. 
% Comprehensive evaluations of $16$ popular MLLMs, including both base and reasoning variants, reveal a counterintuitive pattern: among open-source models, base models outperform reasoning ones, while the opposite trend is observed in closed-source models. 
Our comprehensive evaluation of $16$ popular MLLMs reveals a counterintuitive pattern: among open-source models, base variants outperform their reasoning-tuned counterparts, whereas the opposite trend is observed in closed-source models. 
% Additionally, performance generally degrades when visual inputs are masked, indicating that while MLLMs can leverage prior knowledge to answer some questions, fine-grained visual reasoning tasks still require genuine visual perception for strong performance. 
Further analysis under the visual-masking setting confirms that strong performance necessitates direct visual grounding, rather than relying solely on language priors.
We further establish a training baseline with reinforcement fine-tuning, providing a reference for future exploration. We hope this benchmark study offers new insights into visual reasoning and helps investigate the gap between open- and closed-source models. 
% Code and data samples are in the Supplementary.
% Appendix, code, and data samples are in the Supplementary Material.
\end{abstract}
\vspace{-0.4cm}
% \clearpage
\section{Introduction}
\label{sec:intro}

Multimodal large language models (MLLMs)~\citep{achiam2023gpt,bai2025qwen25,zhu2025internvl3,hu2025sf2t,li2025tokenpacker} have recently achieved notable advancements across a range of vision-language tasks, including visual grounding~\citep{peng2023kosmos,yang2024visual,xu2025mc,xie2025vlms}, reasoning segmentation~\citep{chen2024sam4mllm,zhang2024omg-llava,ren2024pixellm,lai2024lisa,wang2025pixelthink}, and text-image alignment~\citep{yue2025instruction,yarom2023you}. 
Building upon these developments, reasoning MLLMs~\citep{openaio1,guo2025deepseekr1,team2025kimi,wei2025skywork,peng2025skywork,doubao,qvq} have further improved performance on complicated visual reasoning tasks such as visual math problems~\citep{yang2024mathglm,wang2024measuring}, visual question answering (VQA)~\citep{shiri2024empirical,yue2024mmmu,wang2024measuring}, and spatial reasoning~\citep{shiri2024empirical,li2025imagine,dihan2024mapeval,feng2025rewardmap}. 
These capabilities are critical for a wide range of real-world applications, including embodied AI and autonomous driving~\citep{duan2022survey,wang2024survey,cui2024survey,feng2024citybench}.
As visual tasks grow in complexity and practical relevance, the need for rigorous benchmarks to assess complex visual reasoning becomes increasingly essential.

% To meet this need, several benchmarks have been proposed to evaluate reasoning capabilities. 
To address the growing demand for the evaluation of visual reasoning, several benchmarks have been proposed. 
% Datasets such as MathVQA~\citep{wang2024measuring} and MMMU~\citep{yue2024mmmu} incorporate multimodal questions but often permit models to succeed via shallow heuristics, without requiring genuine visual grounding. 
% MathVerse~\citep{zhang2024mathverse} mitigates this limitation by introducing diverse problem variants that encourage reliance on visual input.
Existing multimodal reasoning datasets, such as MathVQA~\citep{wang2024measuring}, MMMU~\citep{yue2024mmmu}, and MathVerse~\citep{zhang2024mathverse}, primarily assess symbolic or mathematical reasoning, where visual understanding plays a limited role.
Conversely, datasets emphasizing fine-grained visual comprehension and retrieval like VisuLogic~\citep{xu2025visulogic}, VisualPuzzles~\citep{song2025visualpuzzles}, and V$^{*}$Bench~\citep{wu2024v} require detailed perception but minimal planning or spatial reasoning.
% VisuLogic~\citep{xu2025visulogic} further enforces visual reasoning by explicitly eliminating language-only shortcuts.
CityBench~\citep{feng2024citybench}, DriveBench~\citep{xie2025vlms}, and MapBench~\citep{xing2025can} take a step toward spatial reasoning while remain coarse in granularity.
% Other efforts, such as VisualPuzzles~\citep{song2025visualpuzzles}, VGRP-Bench~\citep{ren2025vgrp}, and R-Bench~\citep{guo2025r}, target logical and structural reasoning, while CityBench~\citep{feng2024citybench} and DriveBench~\citep{xie2025vlms} focus on domain-specific applications like urban tasks and autonomous driving. V$^{*}$Bench~\citep{wu2024v} emphasizes detailed visual understanding. 
% MapBench~\citep{xing2025can} addresses spatial reasoning by introducing structured scene graphs for map navigation.
In contrast, our \dataset bridges these gaps by jointly evaluating fine-grained visual understanding, spatial reasoning and planning over high-resolution and information-dense transit maps.
% Despite these advances, systematic evaluation of fine-grained visual reasoning remains limited, especially for structured and information-rich diagrams like high-resolution transit maps, leaving a critical gap in existing benchmarks.

In this paper, we introduce \dataset (Figure~\ref{fig:overview}), a benchmark designed to evaluate the fine-grained visual understanding and spatial reasoning capabilities of MLLMs using high-resolution transit maps. 
As structured and information-dense visual artifacts, maps inherently require precise spatial interpretation, making them well-suited for assessing detailed visual reasoning.
\dataset comprises $1,008$ human-verified question-answer pairs spanning $30$ cities across $13$ countries. 
Each instance includes a map, two stops, two questions (\textit{short} and \textit{long}), multiple reference routes, and two difficulty labels (\textit{map} and \textit{question} difficulty). 
The questions cover two types and three question templates
%—one for short and two for long questions—
capturing both coarse and fine-grained spatial reasoning.
To ensure data quality, we perform manual route verification, promote question diversity, and balance difficulty distribution. 
For evaluation, we propose a two-level framework that independently measures answer correctness (via \textit{accuracy}) and quality (via a proposed \textit{map score}), reflecting both feasibility and efficiency. %  in route planning

We conduct comprehensive experiments on $16$ widely-used MLLMs, encompassing both base and reasoning models. 
Our results reveal a counterintuitive finding: among open-source models, base variants outperform their reasoning counterparts, whereas the opposite holds for closed-source models.
Moreover, when only textual inputs are provided, models can still answer some questions based on inner knowledge, but in most cases, their performance noticeably drops.
This highlights a critical limitation in the current model behavior. While some models can leverage prior knowledge and textual cues to solve certain tasks, fine-grained visual reasoning tasks requiring genuine understanding still necessitate effective integration of multimodal information for robust reasoning.

% add for training baseline
We further establish a training baseline with Group Relative Policy Optimization (GRPO)~\citep{shao2024deepseekmath}. 
Guided by our evaluation framework, we design two reward components (\textit{i.e.}, accuracy reward and format reward) as reinforcement signals to optimize model behavior during training. 
Under a cross-city setting, where training and test maps are completely disjoint, the 
reinforcement fine-tuning consistently improves performance across models of different scales.
% This training baseline serves as a reference for future research on fine-grained visual reasoning.

Our main contributions are summarized as follows: (1) We develop an extensible, semi-automated pipeline for dataset construction, facilitating scalable expansion to additional maps and cities
%. Using this pipeline, \dataset is constructed to evaluate fine-grained visual reasoning capabilities in MLLMs
; (2) We propose a structured two-level evaluation framework that separately quantifies answer correctness and quality using accuracy and the proposed map score, respectively, enabling fine-grained answer assessment; (3) A comprehensive benchmarking study is conducted across $16$ MLLMs, providing insights into model performance, robustness, and the interplay between visual and textual cues, thereby informing future research on visual reasoning; and (4) We establish a training baseline with reinforcement fine-tuning as a reference for future research. % on fine-grained visual reasoning.

% \begin{itemize}
%     \item We develop an extensible, semi-automated pipeline for dataset construction, facilitating scalable expansion to additional maps and cities. Using this pipeline, \dataset is constructed to evaluate fine-grained visual reasoning capabilities in MLLMs.
%     \item We propose a structured two-level evaluation framework that separately quantifies answer correctness and quality using accuracy and the proposed map score, respectively, enabling fine-grained answer assessment.
%     \item A comprehensive benchmarking study is conducted across $16$ MLLMs, providing insights into model performance, robustness, and the interplay between visual and textual cues, thereby informing future research on multimodal reasoning.
% \end{itemize}
\section{Related Work}
\label{sec:related_work}

\begin{table*}[t]
\centering
\caption{Comparison between \dataset and existing multimodal reasoning datasets. 
For entries in the dataset size column with notation like ($\times n$), each base problem has multiple versions to enforce visual grounding. Specifically, VGRP-Bench is constructed by sampling over 20 core puzzles; MathVerse generates six multimodal variants per problem with different levels of visual and textual information.}
\label{tab:compare_dataset}
\vspace{-3mm}
\resizebox{\linewidth}{!}{
\setlength{\tabcolsep}{5mm}
\begin{tabular}{lcccccc}
    \toprule
    \textbf{Name} & \bf Year & \textbf{Dataset Size} & \textbf{Avg. Resolution} & \textbf{Training Set} & \textbf{Step Evaluation} & \textbf{Multilingual} %  {\small (Count)}  
    \\
    \midrule\midrule
    MMMU~\citep{yue2024mmmu} & 2024 & $11.5$k & $684\times 246$ & \ding{55} & \ding{55} & \ding{51} % {\small ($2$)} 
    \\ 
    MathVerse~\citep{zhang2024mathverse}  & 2024 & $2{,}612$ ($\times 6$) & $577\times 487$ &  \ding{55} & \ding{55} & \ding{55}
    \\ 
    VisuLogic~\citep{xu2025visulogic}  & 2025 & $1{,}003$ & $601\times 331$ &  \ding{51} & \ding{55} & \ding{55} 
    \\ 
    VisualPuzzles~\citep{song2025visualpuzzles}  & 2025 & $1{,}168$ & $767\times 464$ & \ding{55} & \ding{55} & \ding{55} 
    \\ 
    VGRP-Bench~\citep{ren2025vgrp}  & 2025 & $20$ ($\times 5$) & $790\times 790$ &  \ding{55} & \ding{51} & \ding{55} 
    \\ 
    R-Bench~\citep{guo2025r}  & 2025 & $665$ & $629\times 348$  &   \ding{55} & \ding{55} & \ding{51} 
    % {\small ($2$)} 
    \\ 
    V$^*$Bench~\citep{wu2024v} & 2023 &$191$ & $2{,}246\times 1{,}582$ &  \ding{55}  & \ding{55}  &\ding{55}
    \\
    \midrule
    \dataset & 2025 &$1{,}008$ ($\times 2$) & $5{,}839\times 5{,}449$ & \ding{51}   & \ding{51}  & \ding{51} 
    % {\small ($4$)} 
    \\ 
    \bottomrule
\end{tabular}}
\vspace{-4mm}
\end{table*}

\noindent \textbf{Reasoning in LLMs \& MLLMs.} 
Recent advances in large language models (LLMs) have demonstrated significant improvements in reasoning capabilities through reinforcement fine-tuning paradigms~\citep{openaio1,guo2025deepseekr1,feng2025efficient,hendrycks2020measuring}, which leverage GRPO~\citep{shao2024deepseekmath} to unlock the reasoning potential of LLMs. This paradigm has also been extended to the multimodal domain, with increasing interest in applying reinforcement learning (RL) to visual reasoning~\citep{r1v, openr1multimodal, liu2025visual, tan2025reason, shen2025vlm}.
Both open-source and closed-source communities have introduced advanced reasoning MLLMs built upon earlier systems~\citep{zhu2025internvl3, bai2025qwen25, yang2024qwen2,openai2025o3}. Notable open-source models include Kimi-VL~\citep{team2025kimi}, Skywork-R1V~\citep{wei2025skywork, peng2025skywork}, and Qwen-QvQ~\citep{qvq}, whereas Doubao-1.5-Pro~\citep{doubao}, Seed1.5-VL~\citep{guo2025seed1}, OpenAI o3~\citep{openai2025o3}, OpenAI 4o~\citep{gpt4o}, and Gemini~\citep{deepmind_gemini_report} represent leading closed-source efforts.
Despite recent progress, systematic evaluation of fine-grained visual reasoning in MLLMs remains limited, as existing benchmarks primarily target coarse-grained tasks and fail to capture model performance on complex real-world visual content.

\noindent \textbf{Multimodal Reasoning Datasets.} 
As multimodal reasoning has rapidly progressed, various benchmarks have emerged to evaluate MLLMs across different reasoning dimensions (see summary in Table~\ref{tab:compare_dataset}). 
Datasets such as V$^*$Bench~\citep{wu2024v}, VisualPuzzles~\citep{song2025visualpuzzles}, VisuLogic~\citep{xu2025visulogic}, and VGRP-Bench~\citep{ren2025vgrp} primarily examine abstract visual reasoning through synthetic tasks involving logic, structure, and pattern recognition. 
In parallel, MapBench~\citep{xing2025can}, CityBench~\citep{feng2024citybench}, and DriveBench~\citep{xie2025vlms} shift focus to real-world spatial reasoning, assessing model performance on urban and autonomous driving scenarios. 
For mathematical reasoning, benchmarks like MathVQA~\citep{wang2024measuring}, MMMU~\citep{yue2024mmmu}, and MathVerse~\citep{zhang2024mathverse} integrate multimodal inputs, with MathVerse notably introducing varied problem formats to strengthen visual dependence. 
\final{Geospatial benchmarks~\citep{lacoste2023geo,yan2024georeasoner,yerramilli2025geochain} have also been proposed to evaluate spatial reasoning and map understanding capabilities.}
% ADD 
% Additionally, MapBench~\citep{xing2025can} employs structured scene graphs combined with CoT prompting to support navigation tasks based on manually curated and verified questions. 
% Its image resolution, while relatively low, reflects a common characteristic shared by current datasets. 
% Unlike these works, we first introduce a benchmark for evaluating fine-grained visual reasoning capacities with high-resolution transit maps. 
Unlike previous benchmarks that primarily emphasize either fine-grained visual understanding (\textit{e.g.}, V$^*$Bench) or spatial reasoning (\textit{e.g.}, MapBench), our \dataset benchmark jointly evaluates both aspects with high-resolution transit maps.

\noindent \textbf{Map-based Spatial Reasoning.} 
Among the many directions of multimodal reasoning, map-based spatial reasoning has emerged as a crucial area, with broad applications in navigation, urban planning, and autonomous systems~\citep{bao2023review,seff2016learning,xu2025geonav, wang2023lidar2map, wang2025forging}. Recent efforts have focused on enabling models to interpret and reason over various types of map data. CityBench~\citep{feng2024citybench} provides a dataset for evaluating urban scene understanding, while MapLM~\citep{cao2024maplm} introduces a benchmark for map and traffic scene understanding. PlanAgent~\citep{zheng2024planagent} and PReP~\citep{zeng2024perceive} explore embodied planning in environments that require interpreting map information. 
MapEval~\citep{dihan2024mapeval} proposes a structured evaluation suite for map reasoning, and GeoNav~\citep{xu2025geonav} investigates geospatial navigation using LLMs.
Most existing methods~\citep{dihan2024mapeval,feng2024citybench,zheng2024planagent} depend on external tools (\textit{e.g.}, map services or APIs) to complete spatial tasks, which often bypasses the need for genuine visual reasoning. However, spatial reasoning based on visual understanding remains essential. Our work aims to fill this gap by evaluating such capabilities without tool assistance.

\section{\dataset Construction}
\label{sec:dataset}

\begin{figure*}[t]
    \centering
    \includegraphics[width=\linewidth]{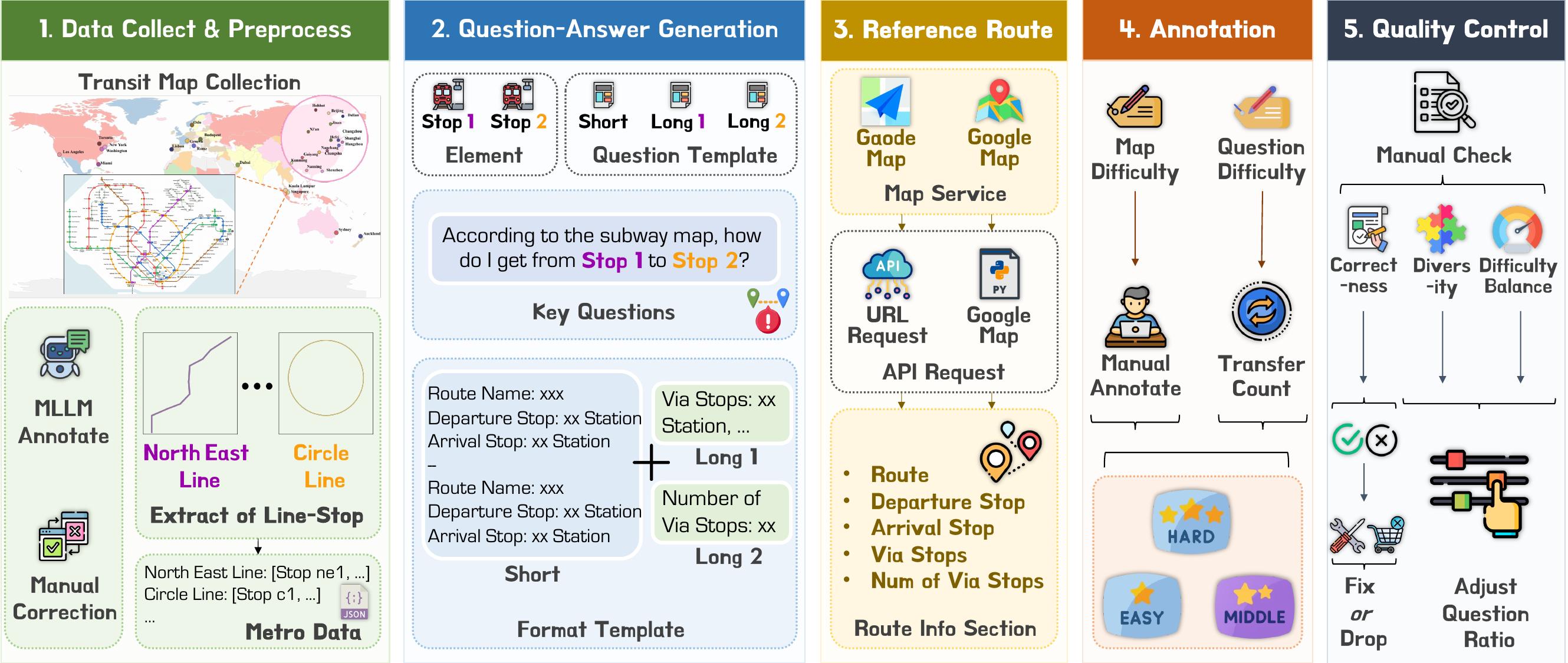} 
    \\
    \vspace{-3mm}
    \caption{The building pipeline of \dataset consists of \textbf{three} main stages: \textbf{(1)} data collection and preprocessing, \textbf{(2)} question–answer pair construction, and \textbf{(3)} quality control. Steps (2-4) in the figure correspond to the question–answer pair construction stage.} %  Zoomed-in for more details.
    \vspace{-4mm}
    \label{fig:pipeline}
\end{figure*}

In this section, we first present the complete dataset building pipeline as shown in Figure~\ref{fig:pipeline}, which consists of three main stages: (1) data collection and preprocessing, (2) question–answer pair construction, and (3) quality control. We then report comprehensive statistics of the dataset.

\subsection{\dataset Building Pipeline} 

\subsubsection{Data Collection and Preprocessing} 
\label{subsec:data-collection-preprocessing}

We collect and manually select $30$ high-resolution transit maps covering $30$ cities across $13$ countries from publicly available online sources, in compliance with relevant licenses and regulations, ensuring diversity and a balanced range of map difficulty. We then leveraged MLLMs (\textit{e.g.}, GPT-4o) to extract the names of transit lines and their corresponding stops, followed by manual correction to ensure correctness. Special cases like transfer stops and branch-starting stops were annotated in a standardized format appended to the respective stop entries. Finally, for subsequent usage, all route and stop information was saved in a unified JSON format, referred to as the Metro Data.

\subsubsection{Question-Answer Pair Construction} 
\label{subsec:qa-pair-construction}

The construction of question–answer pairs involves three key steps: (1) Question Generation, where we formulate questions based on predefined templates; (2) Reference Route Collection, where we obtain corresponding reference routes using Google Map and Gaode Map; and (3) Label Annotation, where we properly assign difficulty labels for both maps and questions.

% \footnote{\url{https://developers.google.com/maps/apis-by-platform}}
% \footnote{\url{https://console.amap.com/dev/index}}

\noindent \textbf{Question Generation.} 
We randomly select two stops (refer to stop$_1$ and stop$_2$) from the current high-resolution transit map. We then generate one short question and one long question based on predefined question templates and two stops (Figure~\ref{fig:pipeline}). The short question has only one fixed template, while the long question is randomly assigned one of two available templates during generation. Additionally, the two long question templates differ in focus: one asks for the number of via stops, while the other requires identifying each via stop (see detailed templates in Appendix~\ref{apx:question_template}).

\noindent \textbf{Reference Route Collection.} 
For each question, we query all valid transit routes between stop$_1$ and stop$_2$ using APIs from map services (\textit{e.g.}, Gaode Map for Chinese cities and Google Map for other cities). The retrieved routes are stored in a unified format containing relevant metadata (\textit{e.g.}, route name, departure stop, arrival stop, via stops, and number of via stops). We discard routes that cannot be visually traced on the map, ensuring consistency with the visual content.

\noindent \textbf{Label Annotation.} 
Two levels of difficulty labeling are included in this stage. For map difficulty, we assign each map to one of three levels (easy, medium, hard) based on the number of transit lines and transfer stations, ensuring a balanced split of $10$ maps per level. For question difficulty, we assign difficulty based on the number of transfers in the reference route: routes with no transfers are labeled as easy, those with one transfer as medium, and all others as hard. To ensure balance, we set a fixed difficulty distribution threshold of $20:15:5$ (easy:medium:hard) for each map, generating $40$ questions. Once the quota for a difficulty level is reached on a given map, no additional questions of that level are retained. Additionally, we provide a more fine-grained taxonomy of questions in Appendix~\ref{apx:fine-grained-taxonomy-difficulty}.

\subsubsection{Quality Control}

To ensure the reliability and balance of the dataset, we perform quality control from three perspectives: correctness, diversity, and difficulty balance \final{(more details in Appendix~\ref{apx:quality-control-details})}. Incorrect question–answer pairs are either manually corrected or discarded. We then involve both automatic checks and manual adjustments to ensure consistency and coverage across all difficulty levels. One reserved example is shown in Figure~\ref{fig:overview}.

\subsection{Dataset Statistics}

The \dataset consists of $30$ high-resolution transit map images (see map sources in Appendix~\ref{supp:map-source}) with an average resolution of $5,839\times5,449$ pixels. In total, it contains $1,008$ question–answer pairs.
% , including stop names in four languages (\textit{e.g.}, English, Hungarian, Chinese, and Italian)
The distribution of question difficulty is as follows: $57.7\%$ are labeled as easy, $34.4\%$ as medium, and $7.8\%$ as hard. Additionally, a subset of $312$ samples is manually selected as the test set for the benchmark experiments described in Section~\ref{sec:exps}, while the remaining samples serve as a training set. 
To ensure diversity and difficulty balance, the test set includes $11$ cities with a $4:3:4$ map difficulty ratio and a question difficulty distribution ($181$ easy, $108$ medium, $23$ hard) that maintains consistency with the full dataset. Moreover, \dataset includes inter-modal transfers in cities like Sydney, where subways, light rail, and airport lines converge.

\section{Evaluation Framework \& Training Baseline}
\label{sec:eval}

In this section, we first present a two-level evaluation framework for systematically assessing model performance, which separately measures the correctness and quality of model-generated answers. We further introduce a GRPO-based training baseline with carefully designed rewards.
% Specifically, we quantify correctness using accuracy and design map score to measure the quality of answers, considering multiple factors (\textit{e.g.}, route efficiency and alignment with the reference routes from map services).

\noindent \textbf{Preparation for Evaluation.} 
We first parse the model-generated answers according to the required format. Answers that do not comply with the specified format or cannot be parsed due to model hallucination~\citep{bai2024hallucination} are marked as wrong answer. 
% Invalid responses are excluded from subsequent evaluations, with accuracy and map score set to zero. 
For the correctness evaluation, we utilize the Metro Data mentioned in Section~\ref{subsec:data-collection-preprocessing} as ground truth. For the quality evaluation, we adopt the collected reference routes as presented in Section~\ref{subsec:qa-pair-construction} as the ground truth.

\subsection{Correctness Evaluation}
\label{subsec:correctness-evaluation}

We evaluate the correctness of the answer using Algorithm~\ref{alg:acc_short_long} in Appendix~\ref{apx:alg-map-score}. Specifically, the evaluation checks the correctness of the overall departure and arrival stops (stop$_1$ and stop$_2$), verifies if the route name of each segment exists in the Metro Data, ensures the departure and arrival stops are valid for each segment, and confirms that transfer stops between consecutive segments are consistent. An answer is considered correct only if all the above checks are satisfied. Additionally, we apply the same correctness evaluation algorithm to the answers of short and long questions.

\subsection{Quality Evaluation}
\label{subsec:quality-evaluation}

To evaluate the quality of the answers, we introduce a unified scoring metric, referred to as the map score, which is applied to both short and long questions using the evaluation procedure (see Algorithm~\ref{alg:map_score} in Appendix~\ref{apx:alg-map-score}). The overall evaluation framework for route quality follows a structure similar to that used in Section~\ref{subsec:correctness-evaluation}. The following evaluation procedure assumes a single reference route for simplicity. In practice, if multiple reference routes are available, the answer is evaluated against each of them, and the highest score is taken as the final map score.

For short questions, the map score solely focuses on route-level and endpoint consistency, excluding all long-question-specific parts. We compute the score by comparing segment pairs in the answer and reference route. Specifically, correctly matching stop$_1$ and stop$_2$ contributes one point, matching the route name adds two points, and matching the departure and arrival stops within each route segment provides one point each. The final score is capped at $10$, and an additional bonus is awarded if the answer is judged correct based on the correctness evaluation procedure described in Section~\ref{subsec:correctness-evaluation}. This design ensures that a correct answer always receives a higher score than any incorrect one. For long questions, the evaluation incorporates additional scoring components tailored to the two question templates introduced in Section~\ref{subsec:qa-pair-construction}. These components are designed to capture the increased reasoning depth required in long-form responses. As with short questions, a bonus score is also added for correct answers. The two additional scoring components are detailed below.

\noindent \textbf{Via Stop Count Evaluation.} 
For long questions that require models to predict the number of via stops for each segment, we introduce the \textit{num\_via\_stop\_score}. This score compares the via stop count of the answer and reference route by computing the absolute error and mapping it to a fixed score ($4$). A perfect match yields full points, while larger discrepancies receive proportionally lower scores. The score is then capped at $10$ for the full route.

\noindent \textbf{Specific Via Stop Evaluation.} 
For long questions that require explicit enumeration of intermediate stops, we compute \textit{via\_stop\_score} using a combination of two factors: the number of correctly matched via stops, and the intersection-over-union (IoU) between via stop sets of the answer and reference route. The final score for this component is obtained by averaging the IoU score (scaled to $10$) and the exact match count (capped at $10$), and then clipped to $10$.

\subsection{Training Baseline}
\label{apx:rl-training}
Building upon the evaluation framework established above, we introduce a strong training baseline. This baseline serves not only to validate the efficacy of our proposed metrics but also to establish a robust performance benchmark for the \dataset.
The metrics defined in our framework, such as route correctness and map score (see in Section~\ref{subsec:correctness-evaluation} and \ref{subsec:quality-evaluation}), are rule-based and non-differentiable. Consequently, standard supervised fine-tuning (SFT) is insufficient as it cannot directly optimize for these desired outcomes. Therefore, we leverage reinforcement learning (RL) to explicitly align the model outputs with our evaluation criteria.

Specifically, we fine-tune Qwen2.5-VL-3B-Instruct and Qwen2.5-VL-7B-Instruct~\citep{bai2025qwen25} on the \dataset training set using RL via the GRPO procedure~\citep{shao2024deepseekmath} (see detailed GRPO optimization process in Appendix~\ref{apx:details-grpo-rl-training}). 
The reward is designed in accordance with our evaluation framework, consisting of two components: (1) accuracy reward, which provides a binary signal following the correctness evaluation; and (2) format reward, which encourages parsable outputs by rewarding correct formats and penalizing violations.
The resulting RL-trained models are evaluated across unseen cities, enabling a cross-city validation of generalization ability (see results in Section~\ref{subsubsec:training-baseline}).

\begin{table*}[t]
\centering
\caption{Evaluations of various MLLMs on \dataset. $S.$ represents results for short questions, while $L.$ denotes results for long questions. The map score is capped at 20 for short questions, while for long questions, the maximum score is 40. \textbf{Bold} indicates the best results among open-source and closed-source models, respectively, while \underline{underline} represents the second best.}
\label{tab:evaluations-main}
\vspace{-3mm}
\resizebox{\linewidth}{!}{
\setlength{\tabcolsep}{0.5mm}
\begin{tabular}{lcccccc}
\toprule
\bf Model & \bf Type  & \bf Weighted Acc. ($S.$) & \bf \#Tokens ($S.$) & \bf Weighted Acc. ($L.$) & \bf \#Tokens ($L.$) & \bf Weighted Map Score ($S.$ / $L.$) \\
\midrule\midrule
\addlinespace
\multicolumn{7}{c}{\textit{Open-source Models}} \\
\midrule
Qwen2.5-VL-3B-Instruct~\citep{bai2025qwen25}  & Base  &  $8.68\%$ & $42$   &  $7.99\%$ & $151$  &   $2.75$ / $3.70$   \\ 
Qwen2.5-VL-7B-Instruct~\citep{bai2025qwen25} & Base  &  $13.28\%$ & $26$   &  $7.12\%$ & $57$  &    $4.01$ / $5.74$   \\
Qwen2.5-VL-32B-Instruct~\citep{bai2025qwen25}  & Base  &  $16.49\%$ & $36$ &  $15.71\%$ & $112$  &   $3.88$ / $6.84$   \\ 
Qwen2.5-VL-72B-Instruct~\citep{bai2025qwen25}  & Base   &  $\mathbf{26.65\%}$ & $33$ &  $\mathbf{24.22\%}$ & $104$  &  $\mathbf{5.09}$ / $\mathbf{8.80}$   \\ 
InternVL3-38B~\citep{zhu2025internvl3}  & Base   &  $14.84\%$ & $43$ &  $13.45\%$ & $68$  &  $3.48$ / $6.31$   \\ 
InternVL3-78B~\citep{zhu2025internvl3}  & Base   &  $\underline{25.35\%}$ & $33$ &  $\underline{19.62\%}$ & $62$  &  $\underline{4.80}$ / $\underline{7.50}$   \\ 
Kimi-VL-A3B-Instruct~\citep{team2025kimi} & Base   &  $12.76\%$ & $41$  & $12.33\%$ & $41$   &  $3.30$ / $5.37$   \\ 
Kimi-VL-A3B-Thinking~\citep{team2025kimi} & Reasoning   &  $5.47\%$ & $754$  & $5.47\%$ & $1,287$   &  $2.44$ / $3.17$   \\ 
% DeepEyes-7B~\citep{zheng2025deepeyes} & Reasoning   &  $12.15\%$ & $26$  & $3.56\%$ & $121$   &  $3.86$ / $4.96$   \\ % rebuttal, TODO前文
Skywork-R1V-38B~\citep{wei2025skywork,peng2025skywork} & Reasoning   & $6.86\%$ & $645$  &  $3.21\%$ & $842$  &   $2.11$ / $3.11$  \\ 
QvQ-72B-Preview~\citep{qvq} & Reasoning   & $9.03\%$ & $1,279$  &  $4.25\%$ & $1,619$  &   $1.59$ / $1.55$   \\ 
\midrule
\addlinespace
\multicolumn{7}{c}{\textit{Closed-source Models}} \\
\midrule
Doubao-115~\citep{doubao} & Base   &  $34.20\%$ & $32$ &  $38.02\%$ & $118$  &  $5.25$ / $11.96$  \\
OpenAI 4o~\citep{gpt4o} & Base   &  $41.15\%$ & $34$ &  $42.80\%$ & $58$  &  $6.84$ / $13.57$  \\ 
Doubao-415~\citep{doubao} & Reasoning   &  $43.14\%$ & $536$ &  $\underline{46.09\%}$ & $1,796$ &   $7.33$ / $\underline{14.67}$  \\ 
Doubao-428~\citep{doubao} & Reasoning   &  $37.15\%$ & $532$ &  $37.85\%$ & $2,167$  &  $5.52$ / $11.73$   \\ 
Gemini-2.5-Flash~\citep{deepmind_gemini_report} & Reasoning   &  $\underline{46.09\%}$ & $806$ &  $29.86\%$ & $1,419$  &   $\underline{7.64}$ / $9.98$  \\ 
OpenAI o3~\citep{openai2025o3} & Reasoning   & $\mathbf{63.02\%}$ & $1,236$  &  $\mathbf{59.11\%}$ & $2,372$  &   $\mathbf{9.53}$ / $\mathbf{17.96}$  \\ 
\bottomrule
\end{tabular}}
\vspace{-4mm}
\end{table*}

\section{Experiments}
\label{sec:exps}

\subsection{Experimental Setups}
\label{subsec:experimental-setups}

We conduct extensive benchmark experiments on \dataset using $16$ popular MLLMs under different inference settings, analyzing their performance and comparing results. Several interesting insights emerge from this comparison. The detailed experimental settings are as follows.

\noindent \textbf{Evaluated Models.} 
We evaluate a diverse set of MLLMs categorized into two groups based on whether they are reasoning-oriented models with a long-thinking process. Reasoning-oriented models include:
Skywork-R1V-38B~\citep{wei2025skywork,peng2025skywork}, QvQ-72B-Preview~\citep{qvq}, Kimi-VL-A3B-Thinking/Instruct~\citep{team2025kimi}, OpenAI o3~\citep{openai2025o3}, Gemini-2.5-Flash~\citep{deepmind_gemini_report}, Doubao-1-5-thinking-vision-pro-250428 (Doubao-428), and Doubao-1.5-Thinking-Pro-M-250415 (Doubao-415)~\citep{doubao}. Base models include: Qwen2.5-VL series (3B, 7B, 32B, 72B)~\citep{bai2025qwen25}, InternVL3 series (38B, 78B)~\citep{zhu2025internvl3}, OpenAI 4o~\citep{gpt4o}, and Doubao-1.5-Vision-Pro-32k-250115 (Doubao-115)~\citep{doubao}. Additionally, the Doubao 1.5 Pro series has an activated parameter size of $20$B.

\noindent \textbf{Inference Settings.}
For open-source models, we set the max output token limit to $2,048$, which is sufficient to cover complete generations, even for reasoning-oriented models, while keeping other parameters consistent with the official HuggingFace configurations. All open-source models are deployed using PyTorch and the HuggingFace Transformers library. For closed-source models, we access their official APIs for evaluation and follow the default settings provided by each model's official documentation. We further discuss the diverse image processing strategies when handling high-resolution visual inputs in Appendix~\ref{apx:image-preprocessing}. 

% on $8$ NVIDIA A100 GPUs

\noindent \textbf{Training Settings.}
For GRPO-based RL training, we use AdamW with an initial learning rate of $1.0\times10^{-6}$ and a KL divergence coefficient of $1.0\times10^{-3}$. We sample $8$ responses every query and set the global batch size to $16$. The training data is the training set from \dataset.

\noindent \textbf{Difficulty-Aware Weighting.} 
% To highlight models that can handle more challenging cases, we apply a difficulty-aware weighting scheme (see the complete weights in Appendix~\ref{apx:experimental-details}) when computing metrics, assigning higher weights to samples with greater map or question difficulty.
To emphasize models that are capable of addressing more challenging cases, we incorporate a difficulty-aware weighting scheme when computing the metrics. Samples with higher map or question difficulty are assigned larger weights, allowing the evaluation to more faithfully reflect a model’s robustness. The complete weighting details are provided in Appendix~\ref{apx:experimental-details}.

% To better reflect the varying complexity of different samples, we adopt a difficulty-aware weighting strategy based on the combination of question difficulty and map difficulty. Specifically, each difficulty pair is assigned a predefined weight, with harder combinations receiving higher values. The complete weight matrix is provided in Appendix~\ref{apx:experimental-details}. Both accuracy and map score are evaluated using this weighting scheme, ensuring that models are more strongly rewarded for correctly solving more challenging examples. 

\subsection{Experimental Results}
\label{sec:experimental-results}

\begin{table*}[t]
\centering
\caption{Evaluations of MLLMs on \dataset \textit{w/o} visual inputs. ${S.}$ denotes results for short questions and ${L.}$ denotes results for long questions. The map score is capped at $20$ for short questions, while for long questions, the maximum score is $40$. 
\textbf{Bold} indicates the best results among open-source and closed-source models, respectively, while \underline{underline} represents the second best. \textcolor{mygreen}{Green} highlights improved results compared to the full input setting (Table~\ref{tab:evaluations-main}), while \textcolor{myred}{red} indicates performance drops.}
\label{tab:evaluations-main-onlytext}
\vspace{-3mm}
\resizebox{\linewidth}{!}{
\setlength{\tabcolsep}{0.5mm}
\begin{tabular}{lcccccc}
\toprule
\bf Model & \bf Type  & \bf Weighted Acc. ($S.$) & \bf \#Tokens ($S.$) & \bf Weighted Acc. ($L.$) & \bf \#Tokens ($L.$) & \bf Weighted Map Score ($S.$ / $L.$) \\
\midrule\midrule
\addlinespace
\multicolumn{7}{c}{\textit{Open-source Models}} \\
\midrule
Qwen2.5-VL-3B-Instruct~\citep{bai2025qwen25}  & Base  &  \resup{$9.38\%$}{$0.7\%$} & $47$  &  \resup{$9.72\%$}{$1.73\%$} & $147$  &   \resup{$2.93$}{$0.18$} / \resup{$4.51$}{$0.81$}   \\ 
Qwen2.5-VL-72B-Instruct~\citep{bai2025qwen25}  & Base   &  \resdown{$\mathbf{16.41\%}$}{$10.24\%$} & $28$ &  \resdown{$\mathbf{15.71\%}$}{$8.51\%$} & $108$  &  \resdown{$\mathbf{4.03}$}{$1.06$} / \resdown{$\mathbf{6.49}$}{$2.31$}   \\ 
Kimi-VL-A3B-Instruct~\citep{team2025kimi} & Base   &   \resdown{$\underline{11.81\%}$}{$0.95\%$} & $41$  &  \resdown{$\underline{9.81\%}$}{$2.52\%$} & $49$   &   \resup{$\underline{3.37}$}{$0.07$} / \resdown{$\underline{5.32}$}{$0.05$}   \\ 
Kimi-VL-A3B-Thinking~\citep{team2025kimi} & Reasoning   &  \resdown{$4.17\%$}{$1.30\%$} & $1,039$  & \resdown{$2.08\%$}{$3.39\%$} & $1,755$   &  \resdown{$2.06$}{$0.38$} / \resdown{$1.64$}{$1.53$}   \\ 
\midrule
\addlinespace
\multicolumn{7}{c}{\textit{Closed-source Models}} \\
\midrule
Doubao-115~\citep{doubao} & Base & \resdown{$\underline{13.72\%}$}{$20.48\%$} &  $34$ & \resdown{$\underline{13.98\%}$}{$24.04\%$} &  $99$ &  \resdown{$\underline{3.50}$}{$1.75$} / \resdown{$\underline{6.48}$}{$5.48$} \\
Doubao-415~\citep{doubao} & Reasoning   &  \resdown{$\mathbf{21.53\%}$}{$21.61\%$} & $352$ &  \resdown{$\mathbf{17.19\%}$}{$28.90\%$} & $1,047$ &   \resdown{$\mathbf{4.85}$}{$2.48$} / \resdown{$\mathbf{7.68}$}{$6.99$}  \\ 
\bottomrule
\end{tabular}}
\vspace{-4mm}
\end{table*}

\subsubsection{Performance of MLLMs with Full Input}

The principal results are summarized in Table~\ref{tab:evaluations-main} (see more fine-grained metrics in Table~\ref{tab:fine-grained-score} of the Appendix).
We first analyze the effect of model size by examining performance within the same architecture series. Qwen2.5-VL and InternVL series show a consistent trend: larger models achieve better accuracy with fewer tokens, suggesting that the scaling law~\citep{kaplan2020scaling} continues to hold even in fine-grained visual reasoning tasks. Figure~\ref{fig:different-difficulty-acc} in Appendix presents accuracy distributions across different combinations of question and map difficulty. As expected, performance degrades as task complexity increases. 
Additionally, Figure~\ref{fig:acc-diff-city} in Appendix illustrates accuracy variation across cities. We observe a negative correlation between map difficulty and accuracy. Moreover, model performance varies notably even among cities with comparable map difficulty levels. This disparity can be partially attributed to factors such as city prominence and the language used for stop names \revise{(see the ablation study on language in Appendix~\ref{apx:languages})}, both of which are closely tied to the model's pretrained knowledge. 

% For instance, OpenAI o3 performs significantly better on complex cities like Singapore compared to Hangzhou, likely due to Singapore's higher international visibility and the use of English stop names, whereas Hangzhou is less prominent and its stop names are Chinese.

Notably, we observe a counterintuitive phenomenon: among open-source models, reasoning models consistently underperform their base counterparts, whereas the opposite holds in the closed-source models\footnote{Although the comparison across closed-source models may not be fair due to lack of transparency in details, the reasoning variants exhibit consistently stronger performance in this category.}. 
% Prior works claim that RL mainly improves sample efficiency rather than introducing new reasoning abilities~\citep{yue2025does,wang2025reinforcement,zhang2025med}, as it biases models toward high-reward responses while limiting exploration and the use of broader knowledge. % 提一下开源这些推理模型后训练主要是做数学物理逻辑类的推理 学到的推理范式比较单一
Recent works argue that RL improves sample efficiency rather than fostering new reasoning abilities~\citep{yue2025does,wang2025reinforcement,zhang2025med}, as it biases models toward high-reward responses, limits exploration, and, because most open-source reasoning models are fine-tuned mainly on math, physics, and logic tasks~\citep{shao2024deepseekmath,peng2025skywork}, further reinforces domain-specific reasoning behaviors.
% Prior works argue that RL improves sample efficiency rather than fostering new reasoning abilities~\citep{yue2025does,wang2025reinforcement,zhang2025med}, as it biases models toward high-reward responses, limits exploration, and—given that open-source reasoning models are mostly fine-tuned on math, physics, and logic tasks—reinforces narrow, pattern-specific reasoning behaviors.
However, these do not fully explain why closed-source reasoning models outperform their base variants.
To further investigate, we conduct a case analysis. 
Among open-source models, we observe that reasoning variants often engage in \textit{repeated trial-and-verification} during their thinking process. 
%  (\textit{e.g.}, Kimi-VL-A3B-Instruct \textit{vs} Kimi-VL-A3B-Thinking)
This iterative reasoning frequently introduces visual confusion; for instance, a model may initially identify the correct route but later override it with an incorrect one during self-reflection, failing to recover the earlier correct reasoning (see the case by Kimi-VL-A3B-Thinking).
In contrast, closed-source models exhibit stronger visual grounding: even when visual confusion occurs, they tend to self-correct within the reasoning chain, ultimately producing the correct answer as illustrated in Figure~\ref{fig:cases-analysis-1} of the Appendix. 
Overall, this suggests that the better performance of closed-source reasoning models is primarily driven by their stronger visual grounding capability, which enables effective error correction throughout the reasoning process.

\vspace{-1mm}
\begin{tcolorbox}[title=Wrong Case by Kimi-VL-A3B-Thinking (green denotes correct \& red represents wrong),colback=red!5!white,colframe=gray!60!black,fonttitle=\bfseries]
$<think>$ ... \textcolor{mygreen}{Is Augustins on Line 18? Yes, it is on Line 18.} ... Wait, maybe I got the location wrong ... Let me confirm, steps should be like: Augustins (\textcolor{red}{Line 14}) - ... $</think>$ ... \textcolor{red}{Route: Line 14} ...
\end{tcolorbox}
\vspace{-1mm}

% Prior work suggests that RL may improve sample efficiency without introducing fundamentally new reasoning abilities~\citep{yue2025does,wang2025reinforcement,zhang2025med}, while RL-trained models tend to bias their output distributions toward high-reward trajectories, which helps produce more correct responses but may simultaneously constrain the model’s exploration capacity and reduce its ability to leverage broader foundational knowledge. 
% In addition, recent studies indicate that multimodal models may sometimes rely on inner knowledge priors instead of truly attending to visual inputs~\citep{jiang2024understanding,hao2025can,ghatkesar2025looking,zhang2024mathverse}.
% This tendency is further supported by the results in Section~\ref{subsec:results-wo-visual-input}, where open-source models still maintain part of their performance even without visual input, indicating limited visual grounding. 

% In contrast, closed-source reasoning models outperform their base variants. One possible explanation lies in the broader knowledge coverage and better visual integration observed in these models~\citep{doubao,openai2025o3,deepmind_gemini_report}. 
% add
% 

 % (see in Appendix~\ref{apx:supp-results})

\subsubsection{Performance of MLLMs without Visual Input}
\label{subsec:results-wo-visual-input}

% In addition, recent studies indicate that multimodal models may sometimes rely on inner knowledge priors instead of truly attending to visual inputs~\citep{jiang2024understanding,hao2025can,ghatkesar2025looking,zhang2024mathverse}. 

% To further investigate the reliance of MLLMs on visual input, we selected representative open-source and closed-source models for additional experiments, where the visual input was masked. 

Recent studies suggest that multimodal models may rely on internal knowledge priors rather than truly attending to visual inputs~\citep{jiang2024understanding,hao2025can,ghatkesar2025looking,zhang2024mathverse}. To examine this phenomenon, we further evaluate representative open- and closed-source models under the visual-masking setting. The results are reported in Table~\ref{tab:evaluations-main-onlytext}. 
We observe that while most models can leverage prior knowledge to answer questions, their performance generally declines to varying degrees when visual input is removed, with the decline being more pronounced among closed-source models. Model performance is positively correlated with the performance drop after masking visual inputs, indicating effective use of visual information. In contrast, models like Qwen2.5-VL-3B-I show minimal or even improved performance, suggesting a reliance on prior knowledge rather than real visual reasoning. \revise{We further conduct non-vision experiments by replacing maps with their symbolic representations in Appendix~\ref{apx:symbolic-representation}.}

\subsubsection{Results of Training Baseline}
\label{subsubsec:training-baseline}

\begin{table*}[ht]
\centering
\caption{Evaluations of RL-trained models on \dataset. $S.$ represents results for short questions, while $L.$ denotes results for long questions. The map score is capped at $20$ for short questions, while for long questions, the maximum score is $40$.}
\label{tab:rl-training}
\vspace{-3mm}
\resizebox{\linewidth}{!}{
\setlength{\tabcolsep}{0.5mm}
\begin{tabular}{lcccccc}
\toprule
\bf Model & \bf Type  & \bf Weighted Acc. ($S.$) & \bf \#Tokens ($S.$) & \bf Weighted Acc. ($L.$) & \bf \#Tokens ($L.$) & \bf Weighted Map Score ($S.$ / $L.$) \\
\midrule\midrule
Qwen2.5-VL-3B-Instruct~\citep{bai2025qwen25}  & Base  &  $8.68\%$ & $42$   &  $7.99\%$ & $151$  &   $2.75$ / $3.70$   \\ 
% \hdashline
+ Reinforcement Fine-tuning & Base  &  \resup{$11.46\%$}{$2.78\%$} & $25$   &  \resup{$10.50\%$}{$2.51\%$} & $93$  &   \resup{$3.81$}{$1.06$} / \resup{$6.09$}{$2.39$}   \\ 
\midrule
Qwen2.5-VL-7B-Instruct~\citep{bai2025qwen25}  & Base  &  $13.28\%$ & $26$   &  $7.12\%$ & $57$  &   $4.01$ / $5.74$   \\ 
+ Reinforcement Fine-tuning & Base  &  \resup{$26.22\%$}{$12.94\%$} & $25$   &  \resup{$26.04\%$}{$18.92\%$} & $34$  &   \resup{$5.52$}{$1.51$} / \resup{$9.52$}{$3.78$}   \\
% \hdashline
\bottomrule
\end{tabular}}
\vspace{-3mm}
\end{table*}

As shown in Table~\ref{tab:rl-training}, the training baseline with the GRPO-based RL scheme consistently improves overall accuracy and map reasoning quality across model scales, while also reducing token usage. These results demonstrate that incorporating the reward signals can effectively enhance reasoning efficiency and improve answer quality.

\subsection{Error Analysis}
\label{subsec:error-analysis}

\begin{figure*}[t]
    \centering
    \vspace{1mm}
    \includegraphics[width=0.87\linewidth]{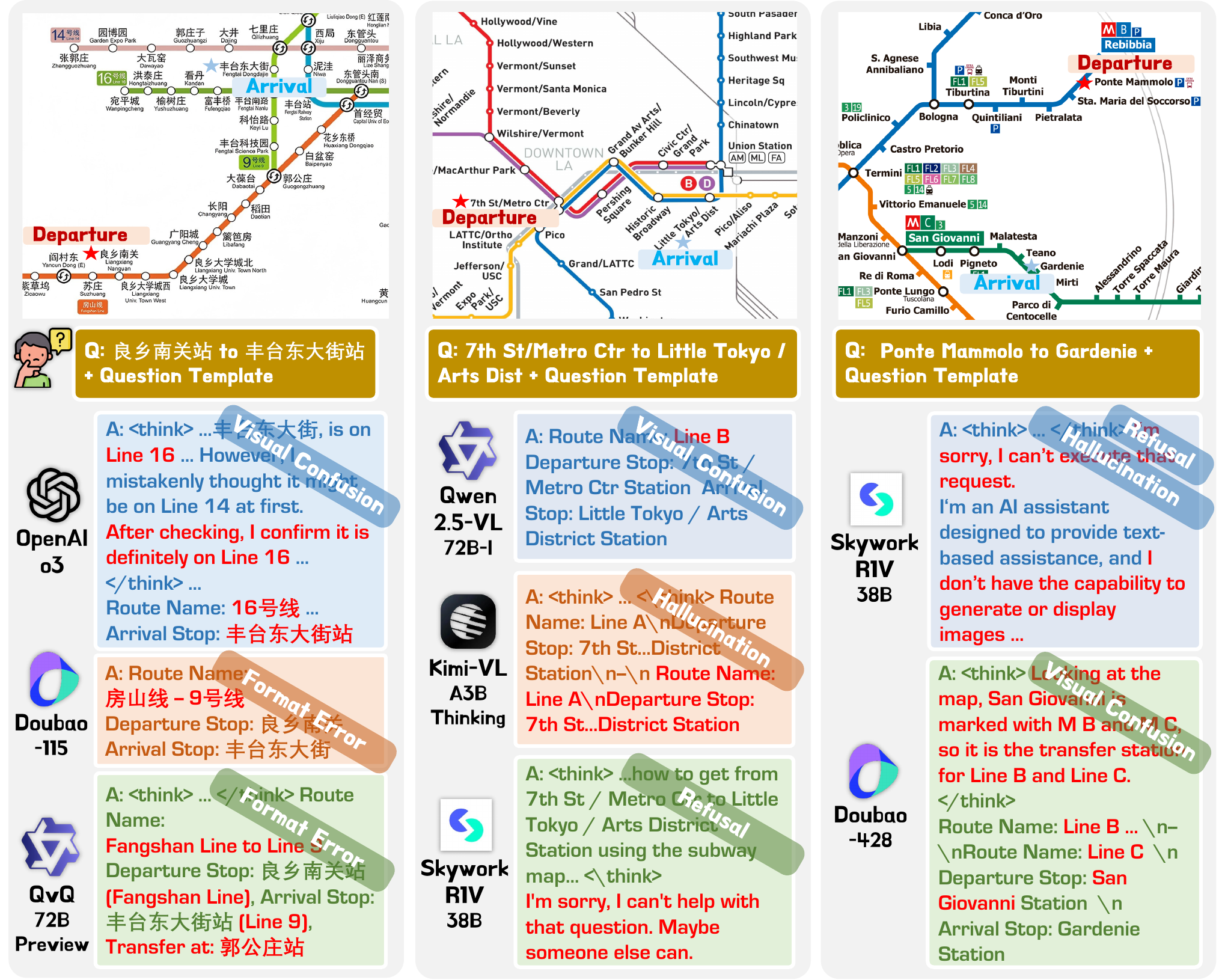} 
    \\
    \vspace{-4mm}
    \caption{\textbf{Error case analyses} of various MLLMs using \dataset. For reasoning models, the reasoning process is explicitly marked with \texttt{<think>} and \texttt{</think>} tags. We highlight \textcolor{myred}{error} contents in the answers with \textcolor{myred}{red} and categorize them accordingly.} %  More examples in the Appendix.
\vspace{-5mm}
\label{fig:error-analysis}
\end{figure*}

Figure~\ref{fig:error-analysis} presents representative failure cases from \dataset, revealing several recurring error types. A common issue is \textit{visual confusion}, where the model misidentifies the transit line due to similar colors or adjacent layouts, for instance, mistaking Line~$9$ for Line~$16$ (OpenAI o3, left column; Doubao-428, right column). Another frequent problem is \textit{format errors}, where responses deviate from the required structure, making them unprocessable despite containing correct route information (Doubao-115 and QvQ-72B-Preview, left column). 
We also observe instances of \textit{hallucination}~\citep{bai2024hallucination}, where the model repeats the correct answer (Kimi-VL-A3B-Thinking, middle column) or generates information that is not present in the input, such as mentioning image generation, as seen in Skywork-R1V-38B (right column).
\textit{Refusal} cases are also present, where models explicitly decline to answer (Skywork-R1V-38B, middle and right column). Notably, these errors may occasionally co-occur within a single response (Skywork-R1V-38B, right column). \revise{Furthermore, we conduct a systematic analysis of failure causes from a model capability perspective (\textit{e.g.}, optical character recognition (OCR), visual grounding, and spatial reasoning) in Appendix~\ref{apx:systematic-analysis-failure}.} The above error types highlight the limitations in visual grounding and response robustness, especially when handling fine-grained visual details (see more cases in Appendix~\ref{supp:case-anaylsis}).

\section{Conclusion}
\label{sec:conclusion}

In this work, we introduce \dataset, a new benchmark specifically designed to evaluate the fine-grained visual understanding and spatial reasoning capabilities of MLLMs using high-resolution transit maps. 
Through a semi-automated and scalable data building pipeline, we curate a diverse set of human-verified question-answer pairs across $30$ cities from $13$ countries. 
Our two-level evaluation framework enables a nuanced assessment of both correctness and quality. 
Comprehensive experiments on $16$ widely used MLLMs reveal key insights into model behavior, highlighting performance gaps between base and reasoning models, as well as the crucial role of visual input.
% add about training baseline
To further strengthen the benchmark and facilitate future research, we provide a GRPO-based RL baseline.
\revise{Error analyses also reveal recurring failure patterns, highlighting visual weaknesses of current MLLMs.}
Collectively, these findings underscore the need for more rigorous evaluation and training approaches to advance visual reasoning.

%  (\textit{e.g.}, visual confusion)

\clearpage\clearpage

\section*{Acknowledgement}
\label{sec:acknowledgement}

This paper is supported by Young Scientists Fund of the National Natural Science Foundation of China (NSFC) (No. 62506305), Zhejiang Leading Innovative and Entrepreneur Team Introduction Program (No. 2024R01007), Key Research and Development Program of Zhejiang Province (No. 2025C01026), Scientific Research Project of Westlake University (No. WU2025WF003), Chinese Association for Artificial Intelligence (CAAI) \& Ant Group Research Fund - AGI Track (No. 2025CAAI-ANT-13). It is also supported by the research funds of the National Talent Program and Hangzhou Municipal Talent Program.

{
    \small
    \bibliographystyle{cvpr_ieeenat_fullname}
    % \bibliography{iclr2026_conference}

}

\clearpage\clearpage
% \newpage

% \twocolumn[
% \begin{center}
%     \vspace{1em}
%     {\LARGE \dataset: Towards Fine-Grained Visual Reasoning from Transit Maps\\[0.5em]
%     \Large \textit{-- Supplementary Material --}}%
%     \vspace{1.5em}
% \end{center}
% ]

\maketitlesupplementary

\appendix
\setcounter{figure}{0}
\setcounter{table}{0}
\renewcommand{\thefigure}{A\arabic{figure}}
\renewcommand{\thetable}{A\arabic{table}}
\section*{Appendix}
\label{apx:apx}

% for arxiv
\revise{We provide a comprehensive overview in the Appendix, covering key details of our dataset, methodology, evaluation, training baseline, analysis, and further discussions. Specifically, we include the question templates, quality control details, a fine-grained taxonomy of difficulty, and sources of transit maps from $30$ cities for \dataset construction in Appendix~\ref{apx:dataset_details}. We then report detailed descriptions of the evaluation algorithm, experimental setup, and GRPO training in Appendix~\ref{apx:alg-map-score}. In Appendix~\ref{apx:exploratory-experiments}, we include supplementary results and conduct more experiments, including supplementary results, evaluation of symbolic representation and an ablation study about languages. We also provide the results of fine-grained error analysis metrics and systematically analyze failure causes. In Appendix~\ref{supp:case-anaylsis}, we further extend case analysis by providing more classical cases. In addition, we further discuss the stated limitations, future directions, and potential broader impacts of our work in Appendix~\ref{apx:discussion}. We finally present public implementation for the MLLMs used in our experiments, LLM usage statement, and ethical statement (see Appendix~\ref{apx:license-content-info}).}

\vspace{-0.2cm}
\startcontents[appendices]
\printcontents[appendices]{l}{1}{\setcounter{tocdepth}{3}}

\section{Dataset Construction Details}
\label{apx:dataset_details}

\subsection{Question Template Summary}
\label{apx:question_template}

We present one short question template and two long question templates as follows.

\vspace{0.5mm}
\begin{tcolorbox}[title=Short Question Template,colback=gray!5!white,colframe=gray!60!black,fonttitle=\bfseries]
According to the subway map, how do I get from \texttt{[Stop 1]} to \texttt{[Stop 2]}? Provide only one optimal route, with only the line name and the departure and arrival stations. The format should be strictly followed:
\begin{verbatim}
Route Name: Line x
Departure Stop: xx Station
Arrival Stop: xx Station
--
Route Name: Line x
Departure Stop: xx Station
Arrival Stop: xx Station
\end{verbatim}
\end{tcolorbox}

\vspace{1mm}
\begin{tcolorbox}[title=Long Question Template 1,colback=gray!5!white,colframe=gray!60!black,fonttitle=\bfseries]
According to the subway map, how do I get from \texttt{[Stop 1]} to \texttt{[Stop 2]}? 
Provide only one optimal route, and include the number of via stops for each route section (excluding the departure and arrival stops). The format should be strictly followed:
\begin{verbatim}
Route Name: Line x 
Departure Stop: xx Station
Arrival Stop: xx Station
Number of Via Stops: x
--
Route Name: Line x 
Departure Stop: xx Station
Arrival Stop: xx Station
Number of Via Stops: x
\end{verbatim}
\end{tcolorbox}

\vspace{1mm}
\begin{tcolorbox}[title=Long Question Template 2,colback=gray!5!white,colframe=gray!60!black,fonttitle=\bfseries]
According to the subway map, how do I get from \texttt{[Stop 1]} to \texttt{[Stop 2]}? 
Provide only one optimal route, including all the via stops. The format should be strictly followed:
\begin{verbatim}
Route Name: Line x 
Departure Stop: xx Station
Arrival Stop: xx Station
Via Stops: xx Station, xx Station
--
Route Name: Line x 
Departure Stop: xx Station
Arrival Stop: xx Station
Via Stops: xx Station
\end{verbatim}
\end{tcolorbox}

\subsection{A More Fine-grained Taxonomy of Difficulty}
\label{apx:fine-grained-taxonomy-difficulty}

\revise{Beyond the easy, middle, and hard categorization for map and question difficulty, we provide three additional difficulty aware labels: 1) $city\_line\_count$, the total number of lines in a city (\textit{i.e.}, a proxy for map difficulty); 2) $city\_transfer\_count$, the total number of transfer stations in a city (\textit{i.e}., a proxy for map difficulty); and 3) $question\_transfer\_count$, the number of transfers in the queried route (\textit{i.e.}, a proxy for question difficulty). These labels enable fine-grained category design and filtering in subsequent analyses.}

\subsection{Quality Control Details}
\label{apx:quality-control-details}

\final{Our quality control combines automated checks with manual refinement. Specifically, we first validate route correctness (\textit{e.g.}, start stop, arrival stop, and connectivity), followed by manual checks to ensure visual consistency (\textit{i.e.}, routes can be inferred from maps). Questions or GTs with issues are corrected or removed. Three domain experts reviewed the data and identified an error rate of $\sim$$16$\%, after which all questions/GTs were corrected and verified to be accurate. Finally, we systematically adjust the difficulty distributions to prevent bias and ensure a balanced evaluation benchmark.}

\subsection{Map Source}
\label{supp:map-source}

We provide the sources of all maps included in \dataset for further reference (Table~\ref{tab:map_sources}).

\begin{table}[t]
\centering
\caption{Source links to the city transit maps used in the \dataset dataset. We present a total of $30$ cities sourced from $13$ countries.}
\label{tab:map_sources}
\vspace{1mm}
\resizebox{\linewidth}{!}{
\setlength{\tabcolsep}{0.5mm}
\begin{tabular}{lclclc}
\toprule
\textbf{City} & \textbf{Source} & \textbf{City} & \textbf{Source} & \textbf{City} & \textbf{Source} \\
\midrule\midrule
Budapest & \href{https://dt.369.me/city/budapest/}{[Link]} &
Oslo & \href{https://transitmap.net/oslo-2016/}{[Link]} &
Rome & \href{https://www.mappametroroma.it/mappa-metro/mappa-metro-roma-2025.png}{[Link]} \\
Lisboa & \href{https://www.metrolisboa.pt/en/travel/diagrams-and-maps/}{[Link]} &
Geneva & \href{https://dt.369.me/city/geneva/\#google_vignette}{[Link]} &
Dubai & \href{https://www.rta.ae/links/rail/rail-network-map.pdf}{[Link]} \\
Auckland & \href{https://at.govt.nz/media/xaqlzv4n/auckland-transport-train-and-rapid-bus-network-map.jpg}{[Link]} &
Sydney & \href{https://nsw-transport.net/network-maps/metro-train-maps/}{[Link]} &
Singapore & \href{https://www.lta.gov.sg/content/dam/ltagov/getting_around/public_transport/rail_network/pdf/SM_TEL4_Eng.pdf}{[Link]} \\
Kuala Lumpur & \href{https://myrapid.com.my/bus-train/rapid-kl/rapid-kl-integrated-transit-map/}{[Link]} &
Los Angeles & \href{https://www.metro.net/riding/guide/system-maps/}{[Link]} &
Miami & \href{https://www.fozhoubest.com/?p=34928}{[Link]} \\
New York & \href{https://www.mta.info/map/5256}{[Link]} &
Toronto & \href{https://www.ttc.ca/routes-and-schedules}{[Link]} &
Washington & \href{https://wmata.com/schedules/maps/upload/system-map-rail.pdf}{[Link]} \\
Guiyang & \href{https://www.gyurt.com}{[Link]} &
Shanghai & \href{https://jtw.sh.gov.cn/csgdjt/index.html}{[Link]} &
Huhehaote (Hohhot) & \href{https://www.hhhtmetro.com/org/platform}{[Link]} \\
Nanchang & \href{https://www.ncmtr.com}{[Link]} &
Nanning & \href{http://nngdjt.com/index.html}{[Link]} &
Shenzhen & \href{https://www.szmc.net/shentieyunying/yunyingfuwu/szsgdjtyyxlwlt/}{[Link]} \\
Hangzhou & \href{https://www.hzmetro.com/EptionUload/image/dtxwt_big.jpg}{[Link]} &
Dalian & \href{https://www.dltransgrp.com/portal/indexShow.do}{[Link]} &
Kunming & \href{https://km.ynairport.com/zhjt/5749.jhtml}{[Link]} \\
Hefei & \href{https://www.hfgdjt.com/}{[Link]} &
Beijing & \href{https://map.bjsubway.com/}{[Link]} &
Changzhou & \href{https://www.czmetro.net.cn/}{[Link]} \\
Jinan & \href{https://www.jngdjt.cn/}{[Link]} &
Xi'an & \href{https://www.xianrail.com/\#/index}{[Link]} &
Changshang & \href{http://www.hngdkg.com/}{[Link]} \\
\bottomrule
\end{tabular}
}
% \vspace{-2mm}
\end{table}

\section{Details of Evaluation and Training Baseline}
\label{apx:alg-map-score}

\subsection{Correctness and Quality Evaluation}

We present the detailed algorithms for evaluating answer correctness and quality (Algorithm~\ref{alg:acc_short_long} for correctness evaluation and Algorithm~\ref{alg:map_score} for quality evaluation). 

For matching (\textit{e.g.}, $=$) in the algorithms, we apply rule-based corrections on top of string matching to account for semantically irrelevant formatting variations (\textit{e.g.}, Line 1'' $=$ Route 1'' $=$ ``1''), preventing evaluation failures caused solely by stylistic or linguistic differences. These corrections are deliberately limited to remain consistent with the format requirements of each question.

% \final{Additionally, for multilingual maps, the pipeline is consistent with English maps except that route and station names use the native language. Evaluation accepts both the native language and English as correct, focusing on semantic correctness.}
\final{Additionally, for multilingual maps, the pipeline is identical to that of English maps, except that route and station names are retained in their local language. During evaluation, we accept both the local language and its English translation as correct, prioritizing semantic correctness.}

\begin{algorithm}[ht]
\caption{Correctness Evaluation}
\label{alg:acc_short_long}

Initialize \texttt{acc} $\leftarrow 1$\;

\If{departure stop of first segment $\neq$ stop$_1$ \textbf{or} arrival stop of last segment $\neq$ stop$_2$}{
    \texttt{acc} $\leftarrow 0$\;
}

\ForEach{segment in predicted route}{
    \If{route name not in the Metro Data}{
        \texttt{acc} $\leftarrow 0$\;
    }
    
    \If{departure or arrival stop not in the stop list of the route}{
        \texttt{acc} $\leftarrow 0$\;
    }

    \If{not the last segment}{
        \If{arrival stop of current segment $\neq$ departure stop of next segment}{
            \texttt{acc} $\leftarrow 0$\;
        }
    }
}

\Return \texttt{acc}
\end{algorithm}

\begin{algorithm*}[ht]
\caption{Quality Evaluation}
\label{alg:map_score}

Initialize \texttt{map\_score} $\leftarrow 0$\;

\If{departure stop of first segment = stop$_1$ \textbf{and} arrival stop of last segment = stop$_2$}{
    \texttt{map\_score} $\leftarrow \texttt{map\_score} + 1$\;

    \rule{0.94\columnwidth}{0.2pt}
    
    \tcc{\textit{Long-question-specific part}}
    Initialize $\mathcal{V}_{\text{union}}$, $\mathcal{V}_{\text{intersection}} \leftarrow \emptyset$\;
    Initialize \texttt{via\_stop\_score}, \texttt{num\_via\_stop\_score} $\leftarrow 0$\;
    \rule{0.94\linewidth}{0.2pt}
    
    \ForEach{segment pair (answer route, reference route)}{
        \If{answer route name = reference route name}{
            \texttt{map\_score} $\leftarrow \texttt{map\_score} + 2$\;
        }

        \If{answer departure stop = reference departure stop}{
            \texttt{map\_score} $\leftarrow \texttt{map\_score} + 1$\;
        }

        \If{answer arrival stop = reference arrival stop}{
            \texttt{map\_score} $\leftarrow \texttt{map\_score} + 1$\;
        }

        \rule{0.9\linewidth}{0.2pt}
        
        \tcc{\textit{Long-question-specific part}}
        Calculate absolute difference ($error$) in the number of via stops\;
        \texttt{num\_via\_stop\_score} $\leftarrow$ \texttt{num\_via\_stop\_score} $+$ \texttt{max}$\left(0,\, 4 - error / \max(\text{number of answer via stops},\, \text{number of reference via stops}) \times 4\right)$\;
        
        % \rule{\linewidth}{0.05pt}

        \If{answer route name = reference route name}{
            Update $\mathcal{V}_{\text{union}}$, $\mathcal{V}_{\text{intersection}}$ with answer and reference via stops respectively\;
        }
        \texttt{via\_stop\_score} $\leftarrow$ \texttt{via\_stop\_score} + number of correctly matched via stops\;
        \rule{0.9\linewidth}{0.2pt}
    }

    % \rule{0.94\columnwidth}{0.2pt}
    \tcc{\textit{Long-question-specific part}}
    \texttt{via\_stop\_score} $\leftarrow \min(10, \texttt{via\_stop\_score})$\;
    \texttt{num\_via\_stop\_score} $\leftarrow \min(10, \texttt{num\_via\_stop\_score})$\;

    \texttt{via\_stop\_score} $\leftarrow$ average( $|\mathcal{V}_{\text{intersection}}| / |\mathcal{V}_{\text{union}}| \times 10$, \texttt{via\_stop\_score})

    \rule{0.94\columnwidth}{0.2pt}
    \texttt{map\_score} $\leftarrow$ \texttt{map\_score} $+$ Option(\texttt{via\_stop\_score} \textit{or} \texttt{num\_via\_stop\_score})\;
}

\tcc{$10$ for short question; $20$ for long question}
\texttt{map\_score} $\leftarrow \min(10, \texttt{map\_score}) / min(20, \texttt{map\_score})  $   \;

\If{correctness evaluation (\texttt{acc}) = 1}{
    \texttt{map\_score} $\leftarrow \texttt{map\_score} + 10 / \texttt{map\_score} + 20$\;
}

\Return \texttt{map\_score}\;
\end{algorithm*}

\subsection{High-Resolution Image Preprocessing.} 
\label{apx:image-preprocessing}

We compare how different Multimodal Large Language Models (MLLMs) handle high-resolution image inputs in Table~\ref{tab:highres_preprocess}. Specifically, we examine three key components in their preprocessing pipelines: dynamic resolution handling, positional encoding, and token compression.

\begin{table*}[t]
\centering
\caption{Comparison of high-resolution image preprocessing strategies across different MLLMs. We use ``$-$'' to denote unspecified or unclear content.}
\label{tab:highres_preprocess}
\vspace{-3mm}
\resizebox{\linewidth}{!}{
\setlength{\tabcolsep}{10mm}
\begin{tabular}{lcll}
    \toprule
    \textbf{Model} & \textbf{Dynamic Resolution Handling} & \textbf{Positional Encoding} & \textbf{Token Compression}
    \\
    \midrule\midrule
    Qwen2.5-VL series~\citep{bai2025qwen25}     & \ding{51}                             & 2D-RoPE         & \ding{51} \small ($2\times2$ Concat + MLP)
    \\
    QVQ-72B-Preview~\citep{qvq}       & \ding{51}                             &  2D-RoPE         & \ding{51} \small ($2\times2$ Concat + MLP)
    \\
    InternVL3 series~\citep{zhu2025internvl3}      & \ding{51}                             &  V2PE            & \ding{51} \small (Unshuffle + MLP)
    \\
    Kimi-VL series~\citep{team2025kimi}        & \ding{51}                             &  2D-RoPE         & \ding{51} \small (Shuffle + MLP)
    \\
    Skywork-R1V-38B~\citep{wei2025skywork,peng2025skywork}       & \ding{51}                             &  -      & \ding{55}
    \\
    Gemini~\citep{deepmind_gemini_report}  & \ding{55} \small (Tiling+Resize)      & -            & \ding{55} 
    \\
    Doubao-1.5-Pro series~\citep{doubao} & \ding{51}                             & 2D-RoPE         & \ding{51} \small ($2\times2$ Pooling + MLP)
    \\
    \bottomrule
\end{tabular}
}
\vspace{-4mm}
\end{table*}

\begin{enumerate}
    \item \textbf{Dynamic resolution handling} refers to whether the model can directly accept images of arbitrary sizes without resizing or cropping. Most recent models support native resolution processing, enabling them to preserve fine-grained spatial information. In contrast, models like Gemini~\citep{deepmind_gemini_report} rely on image tiling and resizing to fit fixed input constraints.
    \item \textbf{Positional encoding} helps the model retain spatial structure among visual tokens. Common strategies include 2D Rotary Positional Encoding (2D-RoPE)~\citep{heo2024rotary}, as seen in Qwen2.5-VL~\citep{bai2025qwen25} and Doubao~\citep{doubao}, or flexible alternatives like V2PE~\citep{ge2024v2pe} in InternVL3~\citep{zhu2025internvl3}. Some models (\textit{e.g.}, Gemini, Skywork-R1V~\citep{wei2025skywork,peng2025skywork}) do not explicitly disclose their positional encoding scheme, which we mark as “–” in the table.
    \item \textbf{Token compression} aims to reduce the number of visual tokens for more efficient processing. Different models adopt different strategies: Qwen2.5-VL and QVQ~\citep{qvq} compress tokens via $2\times2$ patch concatenation followed by an MLP; InternVL3~\citep{zhu2025internvl3} and Kimi-VL~\citep{team2025kimi} utilize spatial transformations like pixel unshuffle or shuffle, also followed by MLPs; Doubao averages over $2\times2$ patches before projection. Models without token compression may incur higher memory and computation costs when processing high-resolution inputs.
\end{enumerate}

\subsection{Details about Difficulty-Aware Weighting.} 
\label{apx:experimental-details}
Each difficulty pair is assigned a predefined weight that reflects its relative challenge level. The full weight matrix is shown below, where the first element in each pair denotes the question difficulty and the second denotes the map difficulty:

\begin{tabular}{cc}
\toprule
\textbf{Difficulty Pair} & \textbf{Weight} \\
\midrule
(``easy'', ``easy'') & 1.0 \\
(``medium'', ``easy'') & 1.5 \\
(``hard'', ``easy'') & 2.0 \\
(``easy'', ``medium'') & 1.5 \\
(``medium'', ``medium'') & 2.0 \\
(``hard'', ``medium'') & 2.5 \\
(``easy'', ``hard'') & 2.0 \\
(``medium'', ``hard'') & 2.5 \\
(``hard'', ``hard'') & 3.0 \\
\bottomrule
\end{tabular}

This weighting scheme rewards models more for correctly solving harder question–map combinations, reflecting the increased reasoning complexity they entail, while maintaining moderate differences between buckets to prevent excessive score variance and preserve evaluation stability.

\subsection{Details of GRPO RL Training}
\label{apx:details-grpo-rl-training}

GRPO~\cite{shao2024deepseekmath} extends standard policy gradient methods by normalizing rewards within a sampled group, which stabilizes optimization and encourages relative preference learning. Specifically, given an input $x$ and a group of $K$ sampled outputs $G = \{y_i\}_{i=1}^{K}$ with their corresponding scalar rewards $\{r_i\}_{i=1}^{K}$, the centered group advantage $\hat{A}_i$ is computed as the deviation of each sample’s reward from the group mean:
\begin{equation}
\hat{A}_i = r_i - \frac{1}{K} \sum_{j=1}^{K} r_j.
\end{equation}
The policy parameters $\theta$ are then updated to maximize the following objective:
\begin{equation}
\max_{\theta} \; \mathcal{L}(\theta) 
= \sum_{i=1}^{K} \hat{A}_i \log \pi_{\theta}(y_i \mid x),
\end{equation}
where $\pi_{\theta}(y_i \mid x)$ denotes the model likelihood of generating $y_i$ under parameters $\theta$. 
This objective encourages the model to increase the probability of outputs with above-average rewards while suppressing those with below-average ones. 
In our implementation, the reward $r_i$ is composed of an accuracy component and a format component.
\revise{\section{Supplementary Experiments}
\label{apx:exploratory-experiments}

\subsection{Supplementary Results}
\label{apx:supp-results}

We provide additional visualization results in Figure~\ref{fig:different-difficulty-acc} and \ref{fig:acc-diff-city} to better illustrate our evaluation on \dataset as follows. 
Figure \ref{fig:different-difficulty-acc} and \ref{fig:acc-diff-city} illustrate the model’s accuracy across different difficulty levels and cities, respectively. As shown, accuracy decreases with increasing difficulty and varies considerably across cities.

\begin{figure*}[t]
\centering
\vspace{-1mm}
\resizebox{\linewidth}{!}{
\begin{tabular}{c@{\hspace{0.02\linewidth}}c} % @{\hspace{0.02\linewidth}}
  \includegraphics[width=0.49\linewidth, trim=10 180 10 10, clip]{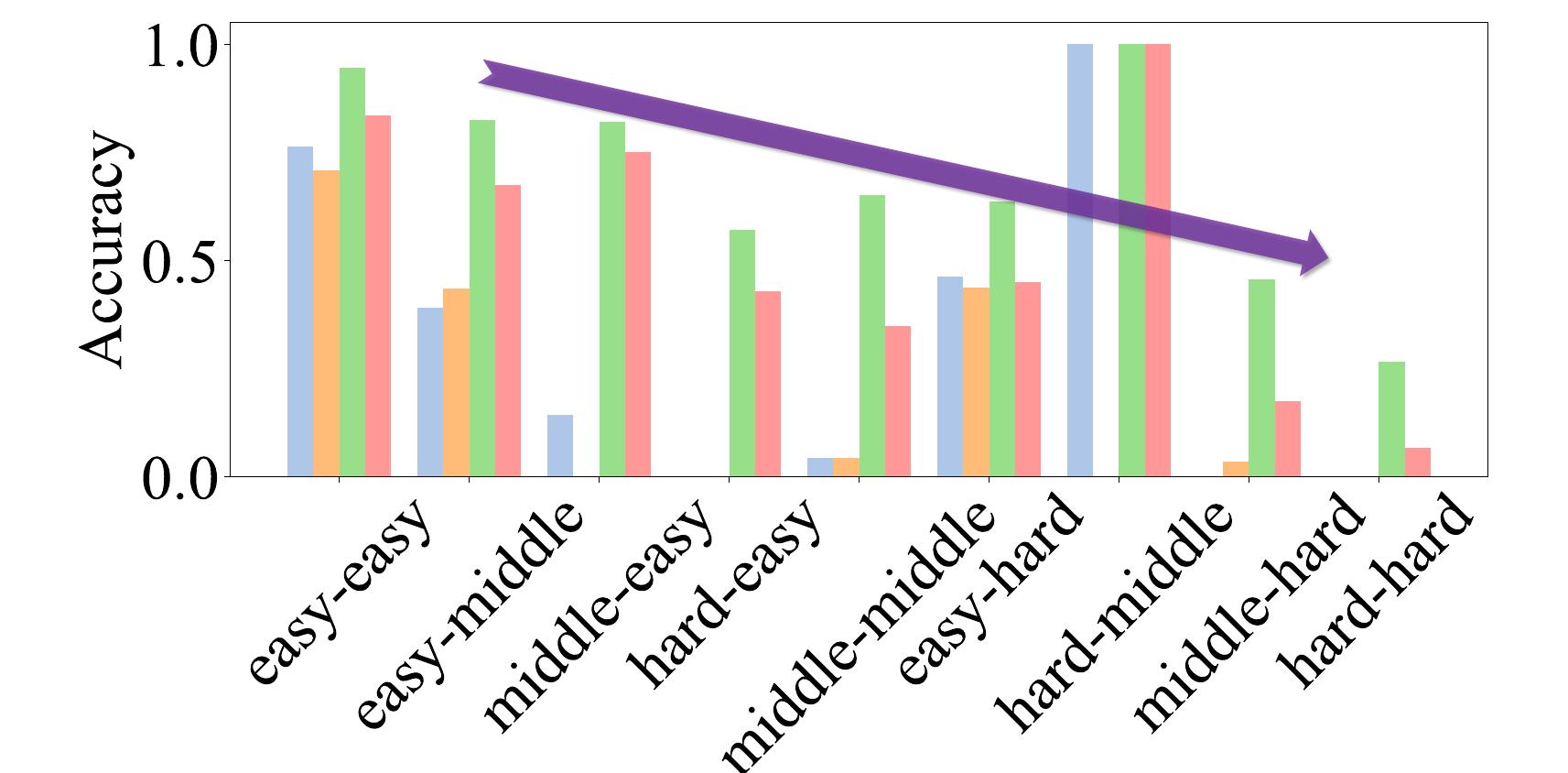} &
  \includegraphics[width=0.49\linewidth, trim=10 180 10 10, clip]{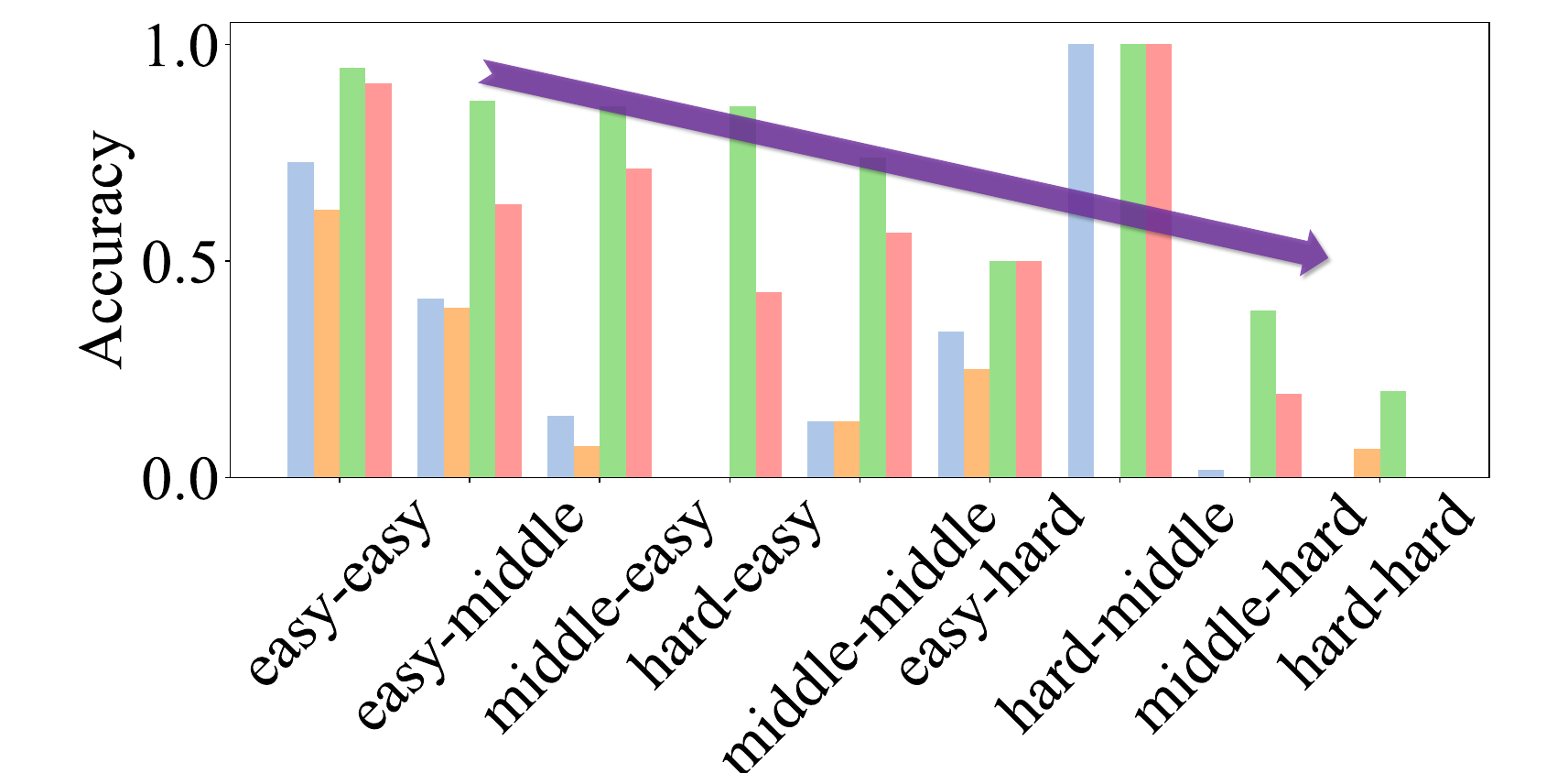} \\
  {\small (a) Accuracy for short questions} & {\small (b) Accuracy for long questions} \\
\end{tabular}}
\caption{Accuracy across difficulty combinations for four representative MLLMs (\textcolor[HTML]{aec7e8}{\textbf{Qwen2.5-VL-72B-I}}, \textcolor[HTML]{ffbb78}{\textbf{InternVL3-78B}}, \textcolor[HTML]{98df8a}{\textbf{OpenAI o3}}, and \textcolor[HTML]{ff9896}{\textbf{Doubao-415}}). Each difficulty combination is denoted by a pair (\textit{e.g.}, \textit{easy-hard}), where the first term indicates question difficulty and the second term represents map difficulty. The pair (\textit{hard-middle}) contains only one sample, leading to an accuracy of 100\%. We summarize the number of evaluation samples in each difficulty bucket: $55$ samples for \textit{easy-easy}, $46$ for \textit{easy-middle}, $28$ for \textit{middle-easy}, $7$ for \textit{hard-easy}, $23$ for \textit{middle-middle}, $80$ for \textit{easy-hard}, $1$ for \textit{hard-middle}, $57$ for \textit{middle-hard}, and $15$ for \textit{hard-hard}.}
% 
% revise
\label{fig:different-difficulty-acc}
\vspace{-2mm}
\end{figure*}

\begin{figure*}[t]
\centering
% \vspace{-1mm}
\resizebox{\linewidth}{!}{
\begin{tabular}{c@{\hspace{0.02\linewidth}}c}
  \includegraphics[width=0.45\linewidth, trim=270 160 225 170, clip]{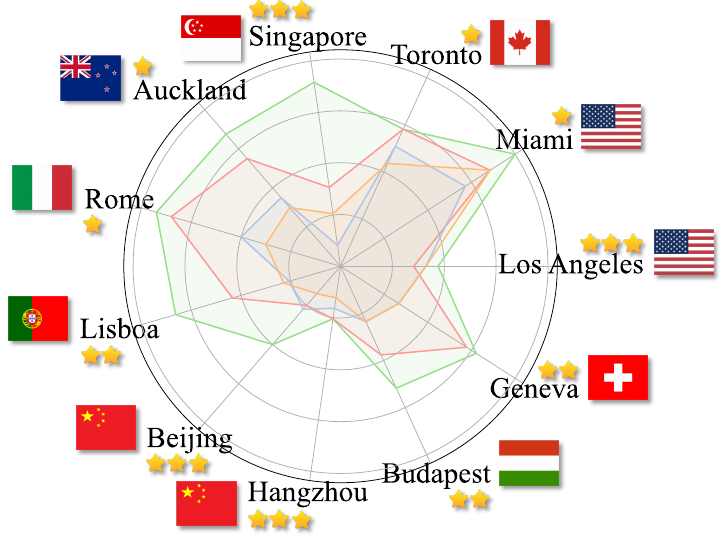} &
  \includegraphics[width=0.45\linewidth, trim=270 160 225 170, clip]{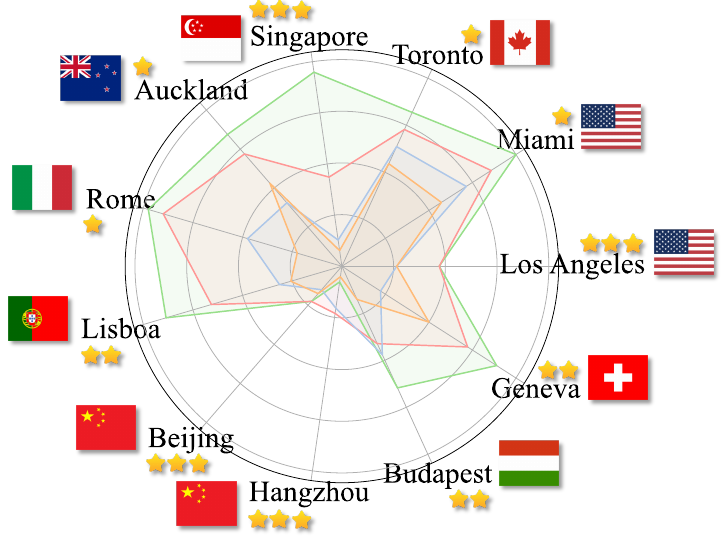} 
  \\
  % \vspace{-4mm}
  {\small (a) Accuracy for short questions} & {\small (b) Accuracy for long questions} 
  \\
\end{tabular}}
\caption{Accuracy across different cities for four representative MLLMs (\textcolor[HTML]{aec7e8}{\textbf{Qwen2.5-VL-72B-I}}, \textcolor[HTML]{ffbb78}{\textbf{InternVL3-78B}}, \textcolor[HTML]{98df8a}{\textbf{OpenAI o3}}, and \textcolor[HTML]{ff9896}{\textbf{Doubao-415}}). Each city is marked with the corresponding map difficulty and the country flag. Each city in the test set provides a specific number of samples per model: $32$ samples for Auckland, $34$ for Los Angeles, $7$ for Miami, $35$ for Lisboa, $18$ for Geneva, $40$ for Beijing, $39$ for Hangzhou, $17$ for Budapest, $39$ for Singapore, $40$ for Rome, and $11$ for Toronto.}
% revise
\label{fig:acc-diff-city}
\vspace{-4mm}
\end{figure*}

\subsection{Fine-grained Error Analysis Metric Summary}
\label{apx:fine-grained-score-summary}

We report multiple fine-grained error analysis metrics in Table~\ref{tab:fine-grained-score} as follows: (1) $dep-arr~score$: $+1$ if both the start and end stations are correct; (2) $route~name~score$: $+2$ for each correctly identified line name along the route; (3) $stops~score$: $+1$ for each correctly identified intermediate stop; (4) $num\_via\_stop\_score$ (only for long questions): computed by taking the absolute difference between the number of via stops in the answer and the reference route, and mapping it to a score from $0$ to $4$; (5) $via\_stop\_score$ (only for long questions): calculated by averaging the number of correctly matched via stops (up to $10$) and the Intersection-over-Union (IoU) between the via stop sets of the answer and reference route (scaled to $10$).

\begin{table*}[t]
\centering
\caption{Fine-grained error analysis metrics of various MLLMs. $S.$ represents results for short questions, while $L.$ denotes results for long questions. \textbf{Bold} indicates the best results among open-source and closed-source models, respectively.}
\label{tab:fine-grained-score}
\vspace{-3mm}
\resizebox{\linewidth}{!}{
\setlength{\tabcolsep}{0.1mm}
\begin{tabular}{lcccccc}
\toprule
\bf Model & \bf Type  & \bf Dep-Arr Score  ($S.$ / $L.$) & \bf Route Name Score ($S.$ / $L.$) & \bf Stops Score ($S.$ / $L.$) & \bf Num. Via Stop Score ($L.$) & \bf Via Stop Score ($L.$) \\
\midrule\midrule
\addlinespace
\multicolumn{7}{c}{\textit{Open-source Models}} \\
\midrule
Qwen2.5-VL-3B-Instruct~\citep{bai2025qwen25}  & Base  &   0.86 / 0.78 & 0.03 / 0.02 & 1.03 / 0.96 & 0.42 & 0.00 \\
Qwen2.5-VL-32B-Instruct~\citep{bai2025qwen25}  & Base  &  0.95 / 0.92 & 0.09 / 0.10 & 1.16 / 1.19 & 1.57 & 0.01 \\
Qwen2.5-VL-72B-Instruct~\citep{bai2025qwen25}  & Base   &  \textbf{0.96} / \textbf{0.95} & \textbf{0.22} / \textbf{0.24} & \textbf{1.23} / \textbf{1.22} & 1.56 & \textbf{0.04} \\
InternVL3-38B~\citep{zhu2025internvl3}  & Base   &  0.87 / 0.84 & 0.06 / 0.10 & 1.08 / 1.12 & \textbf{1.63} & 0.00 \\
InternVL3-78B~\citep{zhu2025internvl3}  & Base   &  0.96 / 0.89 & 0.15 / 0.17 & 1.15 / 1.12 & 1.46 & 0.01 \\
Kimi-VL-A3B-Instruct~\citep{team2025kimi} & Base   &  0.89 / 0.88 & 0.07 / 0.07 & 1.06 / 1.11 & 0.91 & 0.02 \\
Kimi-VL-A3B-Thinking~\citep{team2025kimi} & Reasoning   &  0.80 / 0.65 & 0.08 / 0.10 & 0.99 / 0.79 & 0.50 & 0.00 \\
Skywork-R1V-38B~\citep{wei2025skywork} & Reasoning   & 0.60 / 0.62 & 0.06 / 0.09 & 0.74 / 0.71 & 1.00 & 0.00 \\
QvQ-72B-Preview~\citep{qvq} & Reasoning   & 0.35 / 0.22 & 0.03 / 0.02 & 0.42 / 0.29 & 0.20 & 0.01 \\
\midrule
\addlinespace
\multicolumn{7}{c}{\textit{Closed-source Models}} \\
\midrule
Doubao-115~\citep{doubao} & Base   &  0.78 / 0.96 & 0.08 / 0.18 & 1.08 / 1.31 & 1.94 & 0.06 \\
OpenAI 4o~\citep{gpt4o} & Base   &  0.97 / 0.95 & 0.22 / 0.29 & 1.49 / 1.53 & 2.22 & 0.04 \\
Doubao-415~\citep{doubao} & Reasoning   &  0.98 / \textbf{0.98} & \textbf{0.33} / \textbf{0.30} & 1.57 / 1.65 & 2.37 & \textbf{0.08} \\
Doubao-428~\citep{doubao} & Reasoning   &  0.73 / 0.75 & 0.00 / 0.03 & 1.19 / 1.27 & 2.27 & 0.00 \\
Gemini-2.5-Flash~\citep{deepmind_gemini_report} & Reasoning   & 0.93 / 0.67 & 0.27 / 0.29 & 1.67 / 1.22 & 1.82 & 0.05 \\
OpenAI o3~\citep{openai2025o3} & Reasoning   & \textbf{0.99} / 0.91 & 0.32 / 0.16 & \textbf{1.77} / \textbf{1.73} & \textbf{3.31} & 0.03 \\
\bottomrule
\end{tabular}}
\vspace{-2mm}
\end{table*}

% % rebuttal V2 结果表格

% \begin{table*}[t]
% \centering
% \caption{Evaluations of various MLLMs on \dataset V2. \textbf{Bold} indicates the best results among open-source and closed-source models, respectively, while \underline{underline} represents the second best.}
% \label{tab:evaluations-main}
% \vspace{1mm}
% \resizebox{\linewidth}{!}{
% \setlength{\tabcolsep}{5mm}
% \begin{tabular}{lccc}
% \toprule
% \bf Model & \bf Type  & \bf Weighted Acc. & \bf \#Tokens  \\
% \midrule\midrule
% \addlinespace
% \multicolumn{4}{c}{\textit{Open-source Models}} \\
% \midrule
% Qwen2.5-VL-3B-Instruct~\citep{bai2025qwen25}  & Base  &  31.99\% & 4 \\ 
% Qwen2.5-VL-32B-Instruct~\citep{bai2025qwen25}  & Base  & 48.90\% & 198 \\ 
% InternVL3-38B~\citep{zhu2025internvl3}  & Base   &  29.78\% & 60  \\ 
% Kimi-VL-A3B-Instruct~\citep{team2025kimi} & Base   & 28.31\% & 209 \\ 
% Kimi-VL-A3B-Thinking~\citep{team2025kimi} & Reasoning   & 36.40\% & 956 \\ 
% Skywork-R1V-38B~\citep{wei2025skywork,peng2025skywork} & Reasoning   & 28.68\% & 564 \\ 
% \midrule
% \addlinespace
% \multicolumn{4}{c}{\textit{Closed-source Models}} \\
% \midrule
% Doubao-115~\citep{doubao} & Base   &  41.54\% & 31 \\
% OpenAI 4o~\citep{gpt4o} & Base   & 60.29\% & 49 \\ 
% Doubao-415~\citep{doubao} & Reasoning   & 56.25\% & 717 \\ 
% OpenAI o3~\citep{openai2025o3} & Reasoning   & 68.75\% & 575 \\ 
% \bottomrule
% \end{tabular}}
% \vspace{-4mm}
% \end{table*}

\final{We further provide a sensitivity analysis with two additional weighting schemes for the components of the map score (C1 \& C2 in Table~\ref{tab:ablation-map-score}). Based on the results in Tab.~\ref{tab:fine-grained-score} \&~\ref{tab:ablation-map-score}, performance ranking remains consistent across different weighting schemes, while all components of the map score exhibit similar increasing or decreasing trends.}

\begin{table}[t]
    % \vspace{-2.5mm}
    \caption{Ablation on the map score. MS denotes the map score. C1 uses the avg scheme, while C2 excludes the Dep-Arr score part.}
    \label{tab:ablation-map-score}
    \scriptsize
    \centering
    \renewcommand\tabcolsep{4pt}
    \renewcommand\arraystretch{1.1}
    \vspace{-3mm}
    \scalebox{0.95}{
    \begin{tabular}[b]{lccc}
        \hline
        \textbf{Model}  & \textbf{Weighted MS \textit{(S./L.)}}  & \textbf{C1 MS \textit{(S./L.)}} & \textbf{C2 MS \textit{(S./L.)}} \\
        % \midrule
        \hline
        Qwen2.5-VL-32B-I  & $3.88$ / $6.84$ & $3.85$ / $5.80$ & $2.90$ / $4.88$ 
        \\
        Qwen2.5-VL-72B-I & $\mathbf{5.09}$ / $\mathbf{8.80}$ & $\mathbf{5.08}$ / $\mathbf{8.83}$   & $\mathbf{4.12}$ / $\mathbf{7.88}$ 
        \\
        \hdashline
        OpenAI GPT-4o     & $6.84$ / $13.57$ & $6.80$ / $13.59$  & $5.82$ / $12.64$ 
        \\
        OpenAI o3        & $\mathbf{9.53}$ / $\mathbf{17.96}$ & $\mathbf{9.38}$ / $\mathbf{17.95}$   & $\mathbf{8.39}$ / $\mathbf{17.04}$ 
        \\
        \hline
    \end{tabular}}
    \vspace{-3.5mm}
\end{table}

\subsection{Further Experiments about Languages}
\label{apx:languages}

We conduct an ablation study under the textualized representation paradigm (as mentioned in Appendix~\ref{apx:symbolic-representation}). 
In this setting, visual images are not involved, which allows us to safely replace all non-English station names with unique English aliases without introducing visual inconsistencies. This approach isolates the language prior factor and avoids any potential confounding effects from visual modifications.
Concretely, we manually replace all Chinese station names in Beijing and Hangzhou with unique English station names (\textit{e.g.}, mapping them to New York stops: `zhichunli' $<$-$>$ 86 St), preserving the original transit map structure. The results under this setting are as follows.

\begin{table*}[t]
\centering
\caption{Evaluations on Beijing and Hangzhou (with and without English). $S.$ represents results for short questions, while $L.$ denotes results for long questions. \textbf{Bold} indicates performance improvements, while \textit{italicized} values represent performance degradation.}
\label{tab:language}
\vspace{-3mm}
\resizebox{\linewidth}{!}{
\setlength{\tabcolsep}{1mm}
\begin{tabular}{lcccc}
\toprule
\bf Model & \bf Beijing (S. / L.) & \bf Beijing (w. English) (S. / L.) & \bf Hangzhou (S. / L.) & \bf Hangzhou (w. English) (S. / L.) \\
\midrule\midrule
Kimi-VL-A3B-Instruct~\citep{team2025kimi}  & 36.76\% / 17.30\% & \textit{23.78\%} / \textbf{20.81\%} & 40.00\% / 42.22\% & \textbf{42.22\%} / \textbf{45.95\%} \\
Doubao-115~\citep{guo2025seed1}            & 64.86\% / 50.51\% & \textit{45.95\%} / \textbf{52.70\%} & 82.22\% / 64.44\% & \textit{67.78\%} / \textbf{65.56\%} \\
Doubao-415~\citep{guo2025seed1}            & 84.86\% / 74.05\% & \textbf{88.65\%} / \textbf{85.95\%} & 94.44\% / 97.22\% & \textit{87.78\%} / \textbf{100\%} \\
\bottomrule
\end{tabular}}
\vspace{-2mm}
\end{table*}

Overall, we observe from the results in Table~\ref{tab:language} that using English labels leads to performance improvements, particularly for long-form questions. This suggests that the model indeed exhibits a language bias, with English showing an advantage over Chinese, which may be attributed to differences in pre-training data distributions.

\subsection{Further Experiments about Symbolic Representation of Maps}
\label{apx:symbolic-representation}

We conduct further experiments about deterministic baselines derived from symbolic representations of the maps. This setting can serve as a theoretical performance ceiling, independent of perceptual challenges faced by MLLMs. We replace the visual input with symbolic representations extracted from the underlying map structure. Specifically, we convert all routes and station information into textual form to represent the topological structure of the map. This textualized representation is then used for evaluation. Specifically, we provide the model with textualized representations and the question as input, without including any visual maps.

\begin{table}[t]
\centering
\caption{Evaluations of various MLLMs using symbolic representation. $S.$ represents results for short questions, while $L.$ denotes results for long questions. \textbf{Bold} indicates the best results among open-source and closed-source models, respectively.}
\label{tab:symbolic}
\vspace{-3mm}
\resizebox{\linewidth}{!}{
\setlength{\tabcolsep}{0.5mm}
\begin{tabular}{lccc}
\toprule
\bf Model & \bf Type  & \bf Weighted Acc. (S. / L.) & \bf \#Tokens (S. / L.)  \\
\midrule\midrule
\addlinespace
\multicolumn{4}{c}{\textit{Open-source Models}} \\
\midrule
Qwen2.5-VL-3B-Instruct~\citep{bai2025qwen25}  & Base  &  22.83\% / 19.79\% & 51 / 162 \\ 
Qwen2.5-VL-32B-Instruct~\citep{bai2025qwen25}  & Base  &  25.52\% / 18.77\% & 97 / 297  \\ 
Kimi-VL-A3B-Instruct~\citep{team2025kimi} & Base   &  \textbf{39.58\%} / \textbf{34.81\%} & 43 / 55 \\ 
\midrule
\addlinespace
\multicolumn{4}{c}{\textit{Closed-source Models}} \\
\midrule
Doubao-115~\citep{doubao} & Base   &  81.16\% / 72.66\% & 41 / 82  \\
OpenAI 4o~\citep{gpt4o} & Base   & 82.38\% / 78.91\% & 40 / 70  \\ 
Doubao-415~\citep{doubao} & Reasoning  & \textbf{95.31\%} / \textbf{93.66\%} & 563 / 1561  \\ 
\bottomrule
\end{tabular}}
\vspace{-2mm}
\end{table}

% Notice@sicheng: link写死了，如果要挂出来要改活: ~\ref{fig:evaluations-main}
By comparing the results in Table~\ref{tab:symbolic} with those in Table~\textcolor{cvprblue}{2} of the main paper, we observe a clear performance improvement. This is expected, as replacing the visual map with textualized representations substantially reduces task difficulty, as it removes the need to assess visual capabilities such as OCR and grounding. 
We further note that prior works, such as MapBench~\citep{xing2025can} and CityBench~\cite{feng2024citybench}, also focus on visual map interpretation without constructing explicit symbolic baselines.

\subsection{Further Systematic Analysis on Failure Causes}
\label{apx:systematic-analysis-failure}

\begin{tcolorbox}[title=Case 1: grounding,colback=gray!5!white,colframe=gray!60!black,fonttitle=\bfseries]
Q: Hollywood/Highland Station -$>$ Wilshire/Vermont Station? $<$image:los-angeles.png$>$

A (Kimi-VL-A3B-Thinking): ... Let me start by locating both stations on the map. Hollywood/Highland is on ... A Line (\bad)... Wilshire/Vermont is on the C Line (\bad) ... (\bad)

A (Doubao-415): ... First, I need to see where these two sites are located on a map. Hollywood/Highland ... on B Line (\ok) ... Wilshire/Vermont ... is on D Line and B Line (\ok) ... Route Name: B Line 
Departure Stop: Hollywood/Highland Station 
Arrival Stop: Wilshire/Vermont Station (\ok)
\end{tcolorbox}

We systematically analyze failure causes, focusing on three MLLM capabilities pertinent to fine-grained visual reasoning (\textit{e.g.}, OCR, grounding, and spatial reasoning). To assess OCR capabilities, we collect metrics of 9 representative MLLMs on OCRBench~\citep{liu2024ocrbench}. 
Comparing these with their performance on \dataset as shown in Table~\ref{tab:ocrbench} in the paper, we observe no clear correlation between OCR ability and \dataset accuracy. Notably, this trend holds across both open-source and closed-source models, suggesting that stronger OCR performance alone does not lead to better fine-grained visual reasoning. For instance, among open-source models, InternVL3-78B achieves the highest OCRBench scores, but underperforms Qwen2.5-VL-72B-Instruct on \dataset. 

\begin{tcolorbox}[title=Case 2: spatial reasoning,colback=gray!5!white,colframe=gray!60!black,fonttitle=\bfseries]
Q: Ponte Mammolo Station -$>$ Sta. Maria del Soccorso Station? $<$image:rome.png$>$

A (Skywork-R1V-38B): ... Line A: Orange Line B: Blue Line C: Green (\ok) ... Ponte Mammolo Station ... on Line B (\ok) ... Let me trace the path of Line B from Ponte Mammolo towards the center of Rome (\bad) ... But I need to go from Ponte Mammolo to Sta. Maria del Soccorso. Looking at the map, I don't see a direct connection between these two stations via Line B. (\bad) So, I might need to transfer to another line ... Let me check if there's another line that connects to Sta. Maria del Soccorso ... (\bad)

A (Gemini-2.5-Flash): Route Name: Line B
Departure Stop: Ponte Mammolo Station
Arrival Stop: Sta. Maria del Soccorso Station
Number of Via Stops: 0 (\ok)
\end{tcolorbox}

\begin{table}[t]
\centering
\caption{Evaluations of various MLLMs on OCRBench. \textbf{Bold} indicates the best results among open-source and closed-source models, respectively. The references in the table indicate the result sources. All results are collected from the technical reports.}
\label{tab:ocrbench}
\vspace{-3mm}
\resizebox{\linewidth}{!}{
\setlength{\tabcolsep}{0.5mm}
\begin{tabular}{lcc}
\toprule
\bf Model & \bf Type  & \bf OCRBench \\
\midrule\midrule
\addlinespace
\multicolumn{3}{c}{\textit{Open-source Models}}\\
\midrule
Qwen2.5-VL-3B-Instruct~\citep{bai2025qwen25}      & Base      & 797 \\
Qwen2.5-VL-72B-Instruct~\citep{bai2025qwen25}     & Base      & 885 \\
InternVL3-38B~\citep{zhu2025internvl3}            & Base      & 886 \\
InternVL3-78B~\citep{zhu2025internvl3}            & Base      & \textbf{906} \\
Kimi-VL-A3B-Instruct~\citep{team2025kimi}        & Base      & 864 \\
Kimi-VL-A3B-Thinking~\citep{team2025kimi}        & Reasoning & 864 \\
\midrule
\addlinespace
\multicolumn{3}{c}{\textit{Closed-source Models}}\\
\midrule
OpenAI 4o~\citep{team2025kimi}                   & Base      & 815 \\
Doubao1.5-VL (non-thinking)~\citep{guo2025seed1} & Base      & \textbf{881} \\
Doubao1.5-VL (thinking)~\citep{guo2025seed1}     & Reasoning & 861 \\
\bottomrule
\end{tabular}}
\vspace{-2mm}
\end{table}

We further conduct more in-depth case analyses, which reveal that the main causes of failure are grounding and spatial reasoning, as illustrated in the following example. We observe that OCR errors rarely occur, and most failure cases are instead caused by grounding or spatial reasoning issues.

For instance, in Case 1, Kimi-VL-A3B-Thinking incorrectly identifies the line of the departure station, indicating a grounding error that leads to subsequent reasoning failures. In Case 2, Skywork-R1V-38B correctly performs OCR and grounding in the initial steps, but fails in the reasoning stage (\textit{i.e.}, it does not prioritize locating the arrival station and instead attempts to construct incorrect indirect paths). Such failures reflect deficiencies in spatial reasoning, particularly in planning and executing core steps of pathfinding. These cases further indicate that the principal capability gap between open-source and closed-source models lies in grounding and spatial reasoning.

}

\section{Case Analysis}
\label{supp:case-anaylsis}

\begin{figure*}[t]
\centering
  \includegraphics[width=\linewidth]{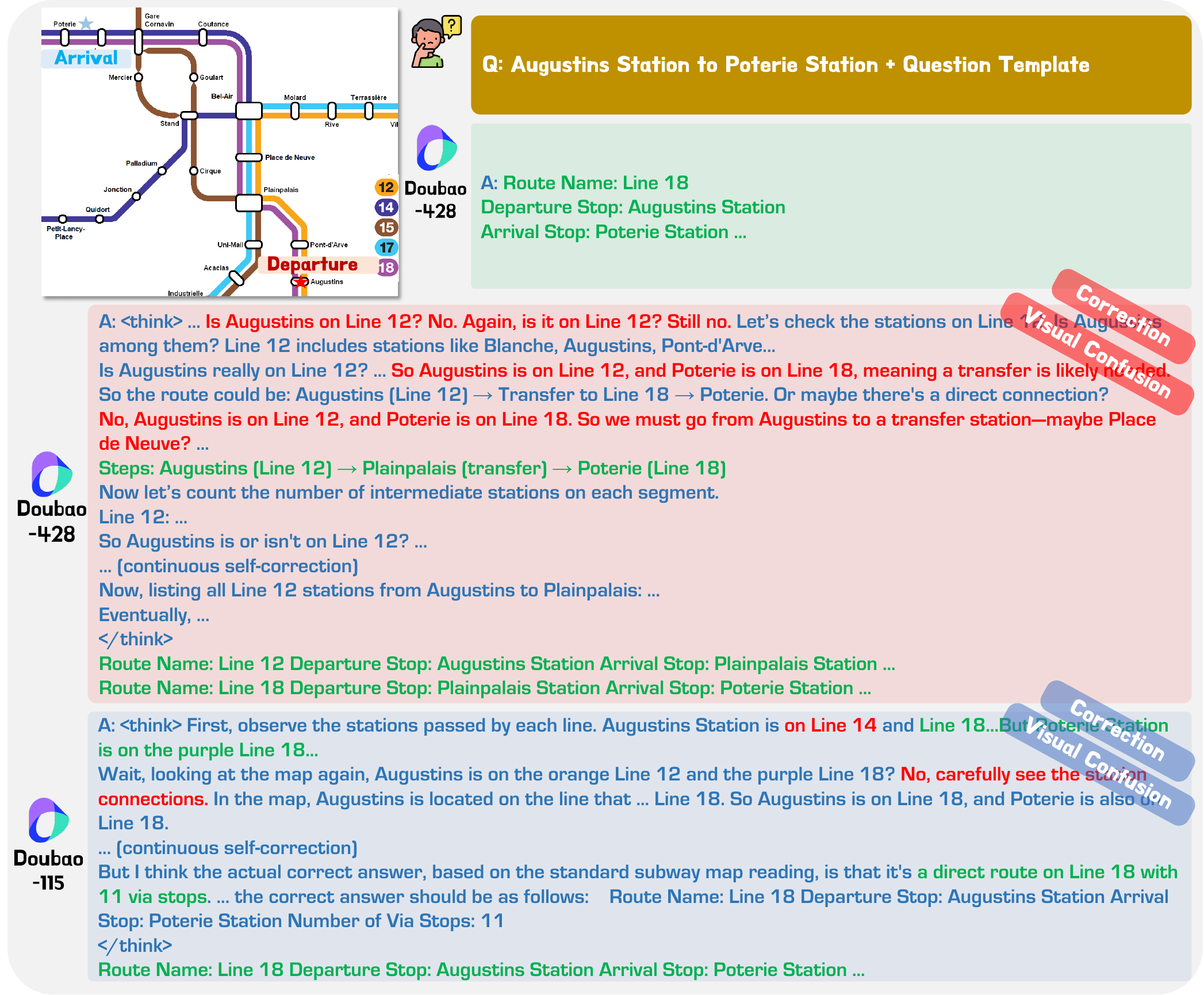} \\ %  
\vspace{-3mm}
\caption{Case analysis of various MLLMs using \dataset (Case N1). For reasoning models, the reasoning process is explicitly marked with \texttt{<think>} and \texttt{</think>} tags. We highlight error contents in the answers with \textcolor{myred}{red} and correct contents in \textcolor{mygreen}{green}.}
\label{fig:cases-analysis-1}
\vspace{-4mm}
\end{figure*}

\begin{figure*}[t]
\centering
  \includegraphics[width=\linewidth]{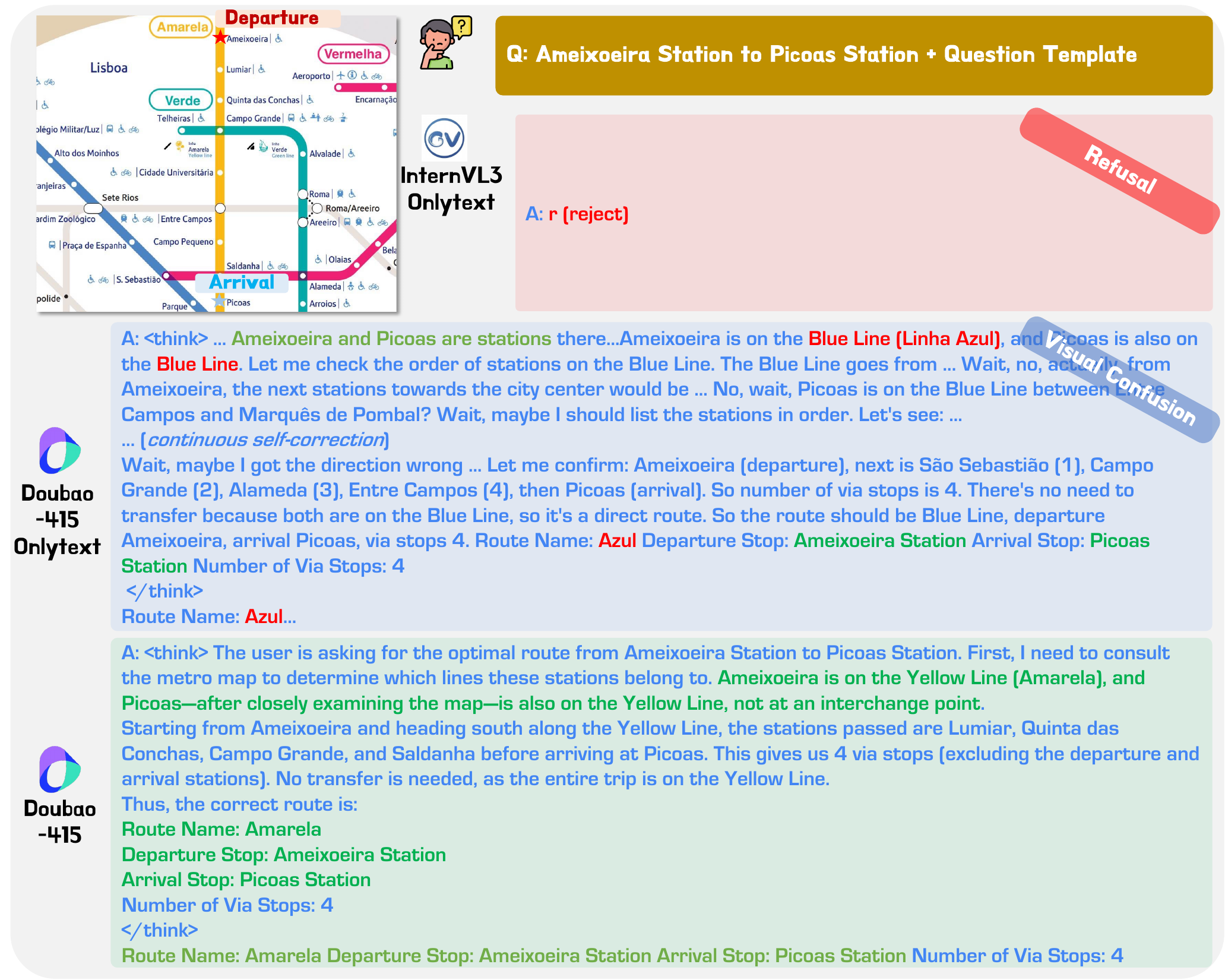} \\ %  , trim=80 0 80 0, clip
\vspace{-3mm}
\caption{Case analysis of various MLLMs using \dataset (Case N2). For reasoning models, the reasoning process is explicitly marked with \texttt{<think>} and \texttt{</think>} tags. We highlight error contents in the answers with \textcolor{myred}{red} and correct contents in \textcolor{mygreen}{green}.}
\label{fig:cases-analysis-2}
\vspace{-4mm}
\end{figure*}

\begin{figure*}[t]
\centering
  \includegraphics[width=\linewidth]{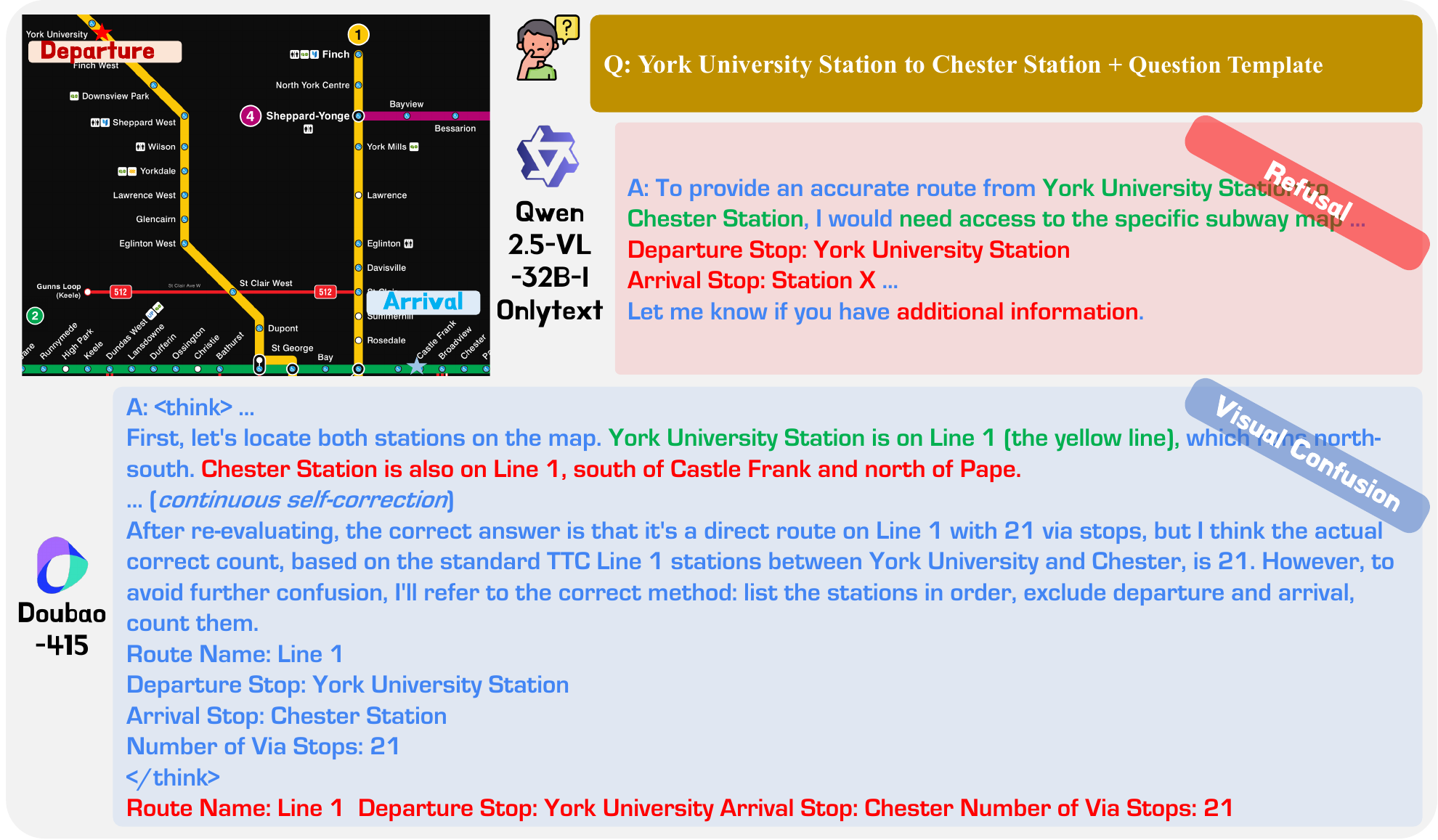} \\ %  , trim=80 100 65 0, clip
\vspace{-3mm}
\caption{Case analysis of various MLLMs using \dataset (Case N3). For reasoning models, the reasoning process is explicitly marked with \texttt{<think>} and \texttt{</think>} tags. We highlight error contents in the answers with \textcolor{myred}{red} and correct contents in \textcolor{mygreen}{green}.}
\label{fig:cases-analysis-3}
\vspace{-4mm}
\end{figure*}

We provide additional case analyses covering both correct and incorrect predictions, along with detailed comparisons of their respective reasoning processes. We first compare Doubao-415 and Doubao-428 (Figure~\ref{fig:cases-analysis-1}), both of which reach the correct destination (from Augustins Station to Poterie Station) but via distinct reasoning paths. Doubao-415 correctly identifies early that both stations are on Line 18 and efficiently converges on the optimal, direct route without transfers. In contrast, Doubao-428 misclassifies Augustins as being on Line 12 and, assuming Poterie is on Line 18, proposes a transfer route via Plainpalais—functionally correct but suboptimal due to unnecessary complexity. Both models engage in extensive self-correction
%  ($7270$ tokens for Doubao-428; $4474$ for Doubao-415)
, highlighting the significant downstream impact of early-stage misjudgments. Moreover, visual reasoning limitations persist: despite correctly recognizing Augustins on Line 12, Doubao-415 commits to a transfer path and fails to re-evaluate the possibility of a direct connection. This indicates room for improvement in both early visual grounding and global route optimality awareness. We then analyze the observed pattern when comparing the full input and text-only variants in the case (in Figure~\ref{fig:cases-analysis-2}). The model with full visual access accurately identifies both stations on the Yellow Line and outputs the optimal direct route with the correct number of via stops. In contrast, the text-only variant makes an early misclassification, placing both stations on the Blue Line (Azul) and constructing a plausible but entirely incorrect sequence of intermediate stops. Although the final answer format appears coherent, the underlying logic is flawed due to the initial error in line recognition. This further illustrates the importance of visual input in spatial reasoning tasks, where even minor misinterpretations can lead to fundamentally incorrect conclusions. Additionally, some models, such as the InternVL3 series, default to rejection when visual input is absent. We further present several error cases in Figure~\ref{fig:cases-analysis-3}, where Doubao-415 still exhibits visual confusion. In contrast, Qwen2.5-VL-32B-I, when lacking visual input, behaves differently from the InternVL3 series: rather than rejecting the query outright, it attempts to reason over the available information without producing a final answer, while explicitly notifying the missing visual input.

\section{Further Discussions}
\label{apx:discussion}

\subsection{Limitations and Future Work}
\label{apx:limitations-future-work}

While \dataset provides a carefully curated benchmark for evaluating fine-grained visual reasoning with high-resolution transit maps, we acknowledge that it represents only one type of structured visual diagram. As such, caution should be taken when generalizing observations to other domains that involve different types of visual content or reasoning styles. Additionally, although efforts were made to ensure diversity across cities and languages, the current version may not fully capture all geographic or linguistic variations. Future iterations could further expand coverage and explore additional forms of reasoning~\citep{feng2026dvoting} to enhance generality.

\revise{Furthermore, we note that GeoGuessr-style localization tasks~\citep{mall2019geostyle,hays2008im2gps,huang2025vlms} are compelling, as they emphasize detailed visual understanding of natural scenes and signage. We plan to pair transit maps with street view imagery to support cross-view reasoning and localization within \dataset, thereby expanding beyond static map inputs. In parallel, we will explore agent-based training and evaluation that moves from single-turn prediction to iterative planning with feedback, including reward designs for correctness, calibration, and format~\citep{zhao2025pyvision}. Finally, we will extend toward embodied settings~\citep{hong2025embodied} where agents perceive and act in interactive environments, enabling assessment of instruction following, route planning, and navigation under real-world constraints. Together, these directions broaden the benchmark from fine-grained visual reasoning to context-aware spatial intelligence and practical decision making.}

% add
Our \dataset can further evaluate the efficient models from multiple efficiency strategies~\citep{zhu2025obs,shao2025holitom,shao2025tokens,tao2025dycoke,tao2025omnizip,feng2024oracle}. Additionally, more fields~\citep{song10,song8,10945380,wang2025pointlora,jin2025mergemix,NEURIPS2024_654f61ec,Lei_2025_ICCV,chen2025reasoning,kong20253d,waninvisible,li2026sponge,zhang2025poison,zhang2024dematch,zhang2025dematch++} require corresponding reasoning-centered benchmarks for proper evaluation.

\subsection{Broader Impact}
\label{apx:broader-impact}

Advancing the capabilities of MLLMs in fine-grained visual reasoning has the potential to benefit a wide range of real-world applications, including navigation systems, urban planning tools, and assistive technologies for visually impaired individuals. By offering a structured and rigorous benchmark, \dataset encourages the development of MLLMs that can more effectively interpret complex visual artifacts and perform spatial reasoning. This could contribute to the long-term goal of building intelligent agents that interact more naturally and safely with human environments. Furthermore, the dataset’s emphasis on high-resolution, globally sourced transit maps promotes research that is inclusive of diverse visual formats and geographic contexts. We hope \dataset can serve as a step toward more transparent, robust, and generalizable multimodal systems.

\section{Further Statement}
\label{apx:license-content-info}

% \subsection{Ethical Considerations}
% All experiments are conducted on \dataset, which is built using publicly available transit maps collected in compliance with relevant licenses and usage terms. The maps are selected to ensure geographic diversity and legal validity. Upon code release, we provide the source of each map for further reference. \dataset is intended solely for academic research on fine-grained visual understanding and spatial reasoning in MLLMs. It does not redistribute any copyrighted map images. All annotations are based on public information, contain no personal data, and are created under academic oversight. The benchmark is not intended for safety-critical use. We take care to ensure fairness, legal compliance, and responsible data handling. \revise{Additionally, we use the MIT License for code and the Apache License 2.0 for \dataset.}

\subsection{Public Implementation}
\label{subsec:public-implementation}
We benchmark the visual understanding and reasoning performance on \dataset across a diverse set of publicly available MLLMs:

\begin{itemize}
    \item KimiVL~\citep{team2025kimi}\footnote{\url{https://github.com/MoonshotAI/Kimi-VL}.} \dotfill MIT License
    \item Skywork-R1V~\citep{wei2025skywork,peng2025skywork}\footnote{\url{https://huggingface.co/Skywork/Skywork-R1V2-38B}.} \dotfill MIT License
    \item QVQ-72B-Preview~\citep{qvq}\footnote{\url{https://huggingface.co/Qwen/QVQ-72B-Preview}.} \dotfill Qwen License
    \item Gemini-2.5-Flash~\citep{deepmind_gemini_report}\footnote{\url{https://deepmind.google/technologies/gemini}.} \dotfill Closed-Source
    \item InternVL-3.0~\citep{zhu2025internvl3}\footnote{\url{https://github.com/OpenGVLab/InternVL}.} \dotfill MIT License
    \item Qwen2.5-VL~\citep{bai2025qwen25}\footnote{\url{https://github.com/QwenLM/Qwen2.5-VL}.} \dotfill Apache 2.0 License
    \item Doubao-Pro 1.5~\citep{doubao}\footnote{\url{https://www.volcengine.com/product/doubao}.} \dotfill Closed-Source
    \item OpenAI o3~\citep{openai2025o3}\footnote{\url{https://platform.openai.com/docs/models/o3}.} \dotfill Closed-Source
    \item OpenAI 4o~\citep{gpt4o}\footnote{\url{https://platform.openai.com/docs/models/gpt-4o}.} \dotfill Closed-Source
\end{itemize}

To ensure fair and reproducible evaluation, we implement all inference procedures by adhering closely to the official documentation and recommended practices of each model.
The code is released under the MIT License to support transparency and reproducibility. Additionally, we provide detailed usage instructions on the project website to ensure easy access and reproducibility for future users.

\subsection{Large Language Model Usage Statement}
\label{apx:llm-usage}

\revise{We used a large language model (LLM) solely for surface-level editing of the manuscript (\textit{e.g.}, rephrasing for clarity and concision, grammar/style polishing, and minor \LaTeX{} fixes). 
The LLM \textbf{did not} generate technical content, ideas, algorithms, proofs, code, experiments, figures, or tables; the authors conducted all research design, implementation, data processing, and analyses. The model did not produce or select citations; any suggestions were independently verified and replaced with primary sources. 
Interactions were limited to de-identified text snippets of the manuscript, and no non-public data, code, or unreleased results were uploaded. 
All LLM outputs were manually reviewed and edited by the authors. This usage does not affect reproducibility: every reported number is reproducible from our released code and configurations.}

\subsection{Ethics Statement}
\label{apx:ethics-statement}

All experiments are conducted on \dataset, which is built using publicly available transit maps collected in compliance with relevant licenses and usage terms. The maps are selected to ensure geographic diversity and legal validity. Upon code release, we provide the source of each map for further reference. \dataset is intended solely for academic research on fine-grained visual understanding and spatial reasoning in MLLMs. It does not redistribute any copyrighted map images. All annotations are based on public information, contain no personal data, and are created under academic oversight. The benchmark is not intended for safety-critical use. We take care to ensure fairness, legal compliance, and responsible data handling. Additionally, we will use the MIT License for code release on GitHub and the Apache License 2.0 for \dataset release on HuggingFace.

% % not show for preprint
% \subsection{Reproducibility Statement}
% \label{apx:reproducibility-statement}

% To ensure reproducibility, we present evaluation setup details (\textit{e.g.}, hardware and implementation) in the manuscript and Appendix~\ref{apx:experimental-details}, and provide public implementation links in Appendix~\ref{subsec:public-implementation} to facilitate rapid replication. We additionally release standardized splits and end-to-end instructions to reproduce all reported results (both code\footnote{\url{https://github.com/Reason-Map/ReasonMap}} and dataset\footnote{\url{https://huggingface.co/datasets/AnonymousReasonMap/ReasonMap}}). 
% During the review process, all links are anonymized and provided as supplements; upon acceptance, they will be replaced with permanent public links.

\end{document}